\documentclass{article}

\usepackage{array}
\usepackage{arxiv}
\usepackage{vcell}
\usepackage[utf8]{inputenc} % allow utf-8 input
\usepackage[T1]{fontenc}    % use 8-bit T1 fonts
\usepackage[colorlinks=true, 
            linkcolor=blue, 
            citecolor=blue, 
            urlcolor=red]{hyperref}
\usepackage{url}            % simple URL typesetting
\usepackage{booktabs}       % professional-quality tables
\usepackage{amsfonts}       % blackboard math symbols
\usepackage{nicefrac}       % compact symbols for 1/2, etc.
\usepackage{microtype}      % microtypography
\usepackage{graphicx}
\usepackage{lipsum}
\usepackage{natbib}
\usepackage{amsmath}
\usepackage{amssymb}
\usepackage{amsthm}         % 只保留这一处
\usepackage{times}
\usepackage{wrapfig}
\usepackage{float}
\usepackage{multirow}
\usepackage{makecell}
\usepackage{algorithm}
\usepackage{algpseudocode}
\usepackage[justification=justified,singlelinecheck=false]{caption} % 删掉上面那个不带选项的 caption
\usepackage{subcaption}
\usepackage{tabularx}
\usepackage{latexsym}
\usepackage{enumitem}
\usepackage{color}
\usepackage{ragged2e}

\newtheorem{thm}{Theorem}
\newtheorem{lem}{Lemma}
\newtheorem{cor}{Corollary}
\newtheorem{defn}{Definition}
\newtheorem{proposition}{Proposition}

\title{Front-door Reducibility: \\ Reducing ADMGs to the Standard Front-door Setting via a Graphical Criterion}

\author{
  Jianqiao Mao \\
  School of Computer Sciencece\\
  University of Birmingham\\
  Birmingham, B15 2TT \\
  \texttt{jxm1417@student.bham.ac.uk} \\
   \And
  Max A. Little \\
  School of Computer Sciencece\\
  University of Birmingham\\
  Birmingham, B15 2TT \\
  \texttt{maxl@mit.edu} \\
}

\renewcommand{\headeright}{}
\renewcommand{\undertitle}{}
\renewcommand{\shorttitle}{\textit{Front-door Reducibility}}

\begin{document}
\maketitle
\begin{abstract}%

Front-door adjustment gives a simple closed-form identification formula under the classical front-door criterion, but its applicability is often viewed as narrow. By contrast, the general ID algorithm can identify many more causal effects in arbitrary graphs, yet typically outputs algebraically complex expressions that are hard to estimate and interpret. We show that many such graphs can in fact be reduced to a standard front-door setting via \emph{front-door reducibility} (FDR), a graphical condition on acyclic directed mixed graphs that aggregates variables into super-nodes $(\boldsymbol{X}^{*},\boldsymbol{Y}^{*},\boldsymbol{M}^{*})$. We characterize the FDR criterion, prove it is equivalent (at the graph level) to the existence of an FDR adjustment, and present FDR-TID, an exact algorithm that finds an admissible FDR triple with correctness, completeness, and finite-termination guarantees. Empirical examples show that many graphs far outside the textbook front-door setting are FDR, yielding simple, estimable adjustments where general ID expressions would be cumbersome. FDR therefore complements existing identification methods by prioritizing interpretability and computational simplicity without sacrificing generality across mixed graphs.

\end{abstract}


\section{Introduction}

\begin{figure}[htbp]
    \begin{centering}
    \includegraphics[width=0.7\linewidth]{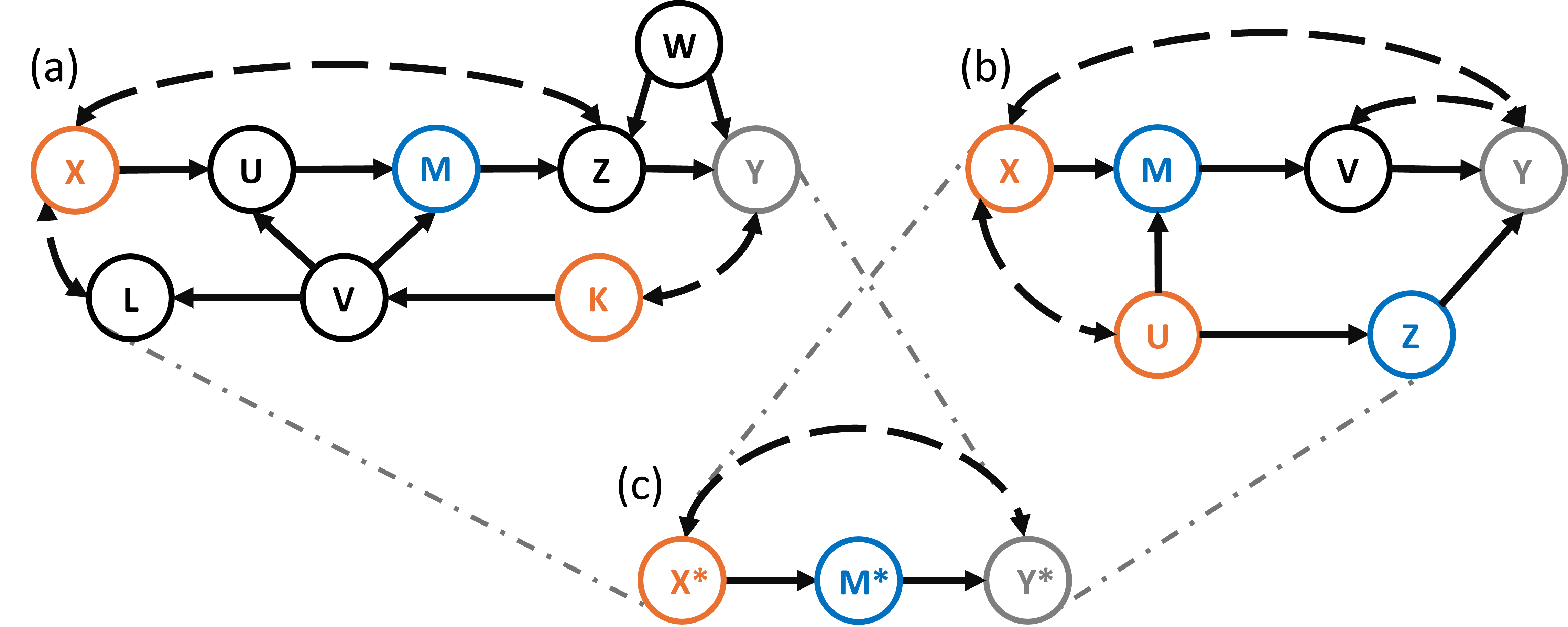}
    \par\end{centering}
    \caption{The example of the equivalence projection from two complicated (but reducible) ADMGs (a) and (b) to a simple, well studied ADMG (c) where ordinary front-door adjustment can be used to obtain the interventional distribution $p\left(\boldsymbol{Y}^{*}\left|do\left(\boldsymbol{X}^{*}\right)\right.\right)$. Different colors show the projection relation of the nodes between (a), (b) and (c), e.g., for (a) $\boldsymbol{X}^{*}=\left\{ X,K\right\} ,$ $\boldsymbol{M}^{*}=\left\{ M\right\} $, and for (b) $\boldsymbol{X}^{*}=\left\{ X,U\right\} $, $\boldsymbol{M}^{*}=\left\{ M,Z\right\} $. \label{fig:complex graph projection}}
\end{figure}

Front-door confounding is a simple but ubiquitous scenario, as shown in Figure~\ref{fig:complex graph projection} (c). The \emph{front-door criterion} was first formally defined by \citet{pearl2009causality}. As shown in Definition~\ref{def: frontdoor}, for a set of variables $\boldsymbol{M}$ that satisfy this criterion, we can use \emph{do}-calculus to compute the interventional distribution \citep{pearl2010introduction}:

\begin{equation}
    p\left(y\left|do\left(x\right)\right.\right)=\int p\left(y\left|x',\boldsymbol{m}\right.\right)p\left(\boldsymbol{m}\left|x\right.\right)p\left(x'\right)d\boldsymbol{m}dx'
    \label{eq: ordinary fd adj formula}
\end{equation}

Despite its familiarity, simplicity and wide applicability \citep{mao2024mechanism}, the ordinary front-door setting is sometimes criticized \citep{robins1995discussion,vanderweele2009relative} for relying on particularly strict structural assumptions as depicted in Definition~\ref{def: frontdoor}. For example, \citet{bellemare2024paper} present a front-door case study in economics with robustness checks on real data. They conclude that the standard front-door adjustment relies on strong structural measurement assumptions and high-quality mediators in empirical studies. However, we argue that the applicability of the front-door criterion is not as limited as it seems: many more complicated causal graphs can be reduced to the front-door criterion. For example, Figure~\ref{fig:complex graph projection} (a) is a complicated ADMG, where the joint causal effect of $\left(X,K\right)$ on $Y$ is theoretically identifiable and its interventional distribution $p\left(y\left|do\left(x\right),do\left(k\right)\right.\right)$ can be identified by the ID algorithm \citep{shpitser2006identification}. Although the ID algorithm is very useful and has been proven to be an effective identification method for an identifiable causal relation in general causal graphs, performing the ID algorithm is not necessarily practical in such a complicated graph, considering the complexity of both the recursive algorithm itself and the difficulty in estimating or using its output expression. In this example, \citeauthor{shpitser2006identification}'s ID algorithm gives the interventional distribution shown as:

\begin{align}
    p\left(y\left|do\left(x\right),do\left(k\right)\right.\right) & =\int p(y\left|x,u,v,m,w,k',z\right.)p(z|u,v,x',m,w,k)p(m,u|v,x,k)\nonumber \\
     & p(x'|w,v,k)p(v|k)p(w)p(k')dx'dmdwdk'dudzdv
    \label{eq:complex id intv prob}
\end{align}

By contrast, the \emph{front-door adjustment} gives a simpler and more practical formula as shown in eq. \eqref{eq: ordinary fd adj formula}. Meanwhile, the complicated ADMG (Figure~\ref{fig:complex graph projection} (a)) can be reduced to an ADMG (Figure~\ref{fig:complex graph projection} (b)) satisfying the front-door criterion with proper operations such as merging and omitting. In this example, the projection can be done by (i) merging $X$ and $K$ as a super-cause node $\boldsymbol{X}^{*}=\left\{ X,K\right\} $; (ii) adjusting on $\boldsymbol{M}^{*}=\left\{ M\right\} $ as the super-mediator node and let $\boldsymbol{Y}^{*}=\left\{ Y\right\} $ be the super-effect node; and (iii) omitting other irrelevant variables. Then the interventional distribution $p\left(y\left|do\left(\boldsymbol{x}^{*}\right)\right.\right)$ is reduced to the ordinary front-door adjustment formula as shown in eq. \eqref{eq: ordinary fd adj formula}. In fact, we show later that a family of ADMGs, e.g., graphs in Figure~\ref{fig:FDR examples} are all reducible to an ordinary front-door setting.

Recent literature has studied and extended the front-door. \citet{fulcher2020robust} propose a generalized front-door approach that aims to identify indirect (mediated) effects even when a direct cause-effect path and unmeasured confounding are present using both parametric and semiparametric methods. However, this method is not applicable for estimating the total effect, and it largely relies on a set of conditions which can be hard to verify in practice. Researchers in the literature \citep{cui2024semiparametric} further extend the framework of proximal causal inference with a hidden mediator variable by \citet{ghassami2021proximal}, deriving influence functions and practical estimators for a broad class of proximal identification problems. However, the reported performance is sensitive to proxy quality and modeling choices. The conditional front-door criterion, an extension of the standard front-door criterion, was proposed by \citet{xu2024causal}, allows an observable confounder acting as the common cause of the cause, mediator and effect variable, while it introduces even stronger conditioning assumptions for the observable confounder. And CFDiVAE is proposed in this work to provide a variational autoencoder-based estimator for identification, whereas it has lower interpretability and incurs a heavier computational burden. More recently, \citet{cakiqi2025algorithmic} provides a new viewpoint from category theory: it turns the identification problem into purely symbolic transformations, offering a formal calculus for implementable identification pipelines. Nevertheless, the empirical performance of the proposed method and its alignment with practical adjustment forms remain to be tested. \citet{wienobst2024linear} develop optimal-time algorithms for finding and enumerating front-door adjustment sets in DAGs under the ordinary front-door criterion, but its scope is still under the classical front-door criterion, whereas we ask when and how more complex ADMGs can be reduced to such front-door forms.

The works above extend front-door based identification in several directions, often relying on strong proximal assumptions or yielding algebraically complex expressions. Our contribution is complementary. We introduce front-door reducibility (FDR), a graphical necessary and sufficient condition that guarantees a simple closed-form FDR adjustment for an enlarged family of ADMGs. We characterize when such an adjustment is valid, and give a complete, sound algorithm (FDR-TID) to identify an admissible FDR triple $(\boldsymbol{X}^{*},\boldsymbol{Y}^{*},\boldsymbol{M}^{*})$. Our equivalence results further show that whenever an FDR adjustment of the form \eqref{eq:fdr adj formula} is universally valid, the underlying ADMG must satisfy the FDR criterion.

\section{Preliminaries}

We first introduce some necessary definitions and theorems in causal inference. In this paper, without special explanations, we use an uppercase letter to denote a variable and a lowercase letter to represent its value. Boldfaced letters are sets of variables and values. Let $\mathcal{G}\left(\boldsymbol{V}\right)$ be an acyclic directed mixed graph (ADMG), where $\boldsymbol{V}$ is the set of nodes indicating observed variables. Denote the ancestor and descendant sets in ADMG $\mathcal{G}$ as $\mathrm{An}_{\mathcal{G}}\left(\cdot\right)$ and $\mathrm{De}_{\mathcal{G}}\left(\cdot\right)$, for example, $\mathrm{An}_{\mathcal{G}}\left(V_{i}\right)$ represents the set of observable variables preceding $V_{i}$ and itself in the topological ordering of $\mathcal{G}$. We represent \emph{m-separation }as $\perp_{m}$. We use $\pi$ to represent a path whose nodes are connected by either directed arcs ($\rightarrow$ or $\leftarrow$) or a bidirected arc ($\leftrightarrow$), i.e., the directed path from $X$ to $Y$ in Figure~\ref{fig:complex graph projection} (c) is $\pi_{X\rightarrow Y}=\left(X\rightarrow M\rightarrow Y\right)$. We use $X\prec_{\mathcal{G}}Y$ to represent that $X$ has smaller topological ordering than $Y$, and $X\preceq_{\mathcal{G}}Y$ if and only if $X\in\mathrm{An}_{\mathcal{G}}\left(Y\right)$. We use $\mathrm{Nb}_{\mathcal{G}}^{\leftrightarrow}\left(\boldsymbol{A}\right):=\left\{ V_{i}\in\boldsymbol{V}\setminus\boldsymbol{A}:\exists A_{i}\in\boldsymbol{A}\text{ s.t. }V_{i}\leftrightarrow A_{i}\text{ in }\mathcal{G}\right\} $ to denote the set of first bidirected-arc-connected neighbours of any node in $\boldsymbol{A}$ in ADMG $\mathcal{G}$. In the following discussion, we use $\mathcal{C}$ as the shorthand for the C-components (Definition~\ref{def: c-component}) of a given ADMG $\mathcal{G}$ without special mention. In this paper, we largely use the three rules of \emph{do}-calculus \citep{pearl2010introduction,pearl2009causality} as shown in Definition~\ref{def: do-calculus rules}. Same as other literature for causal inference, we use $\mathcal{G}_{\overline{X}}$ to denote the modified graph obtained by deleting from $\mathcal{G}$ all arcs incoming to $X$; and use $\mathcal{G}_{\underline{X}}$ to denote the modified graph obtained by deleting from $\mathcal{G}$ all arcs outgoing from $X$. For example, $\mathcal{G}_{\overline{X}\underline{Z}}$ represents a modified graph $\mathcal{G}$ after deleting all incoming arcs to $X$ and outgoing arcs from $Z$.

\section{Front-door reducibility}

\begin{figure}[htbp]
    \begin{centering}
    \includegraphics[width=0.9\linewidth]{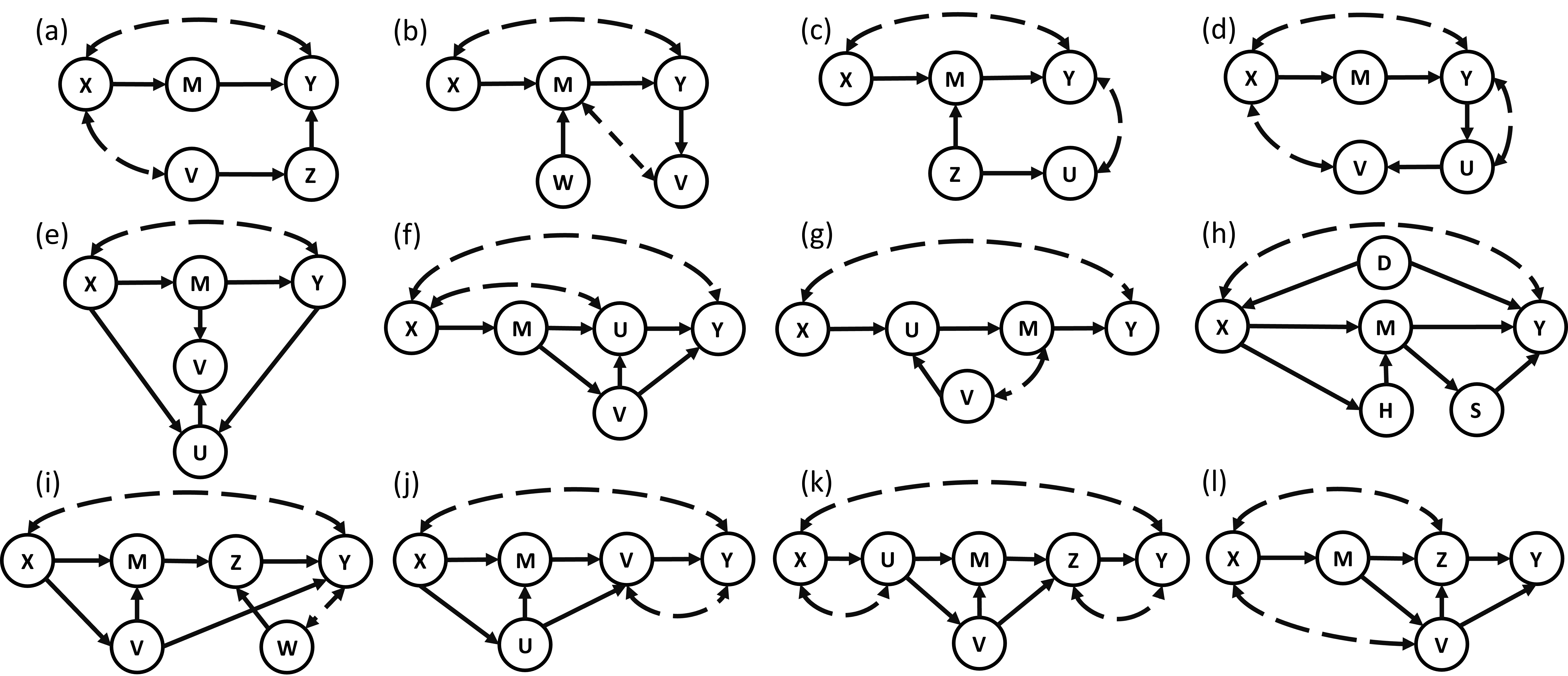}
    \par\end{centering}
    \caption{Example ADMGs relative to the cause and effect of interest pair $\left(X,Y\right)$ satisfying FDR criterion. In (a-h) and (l), an admissible super-cause node is $\boldsymbol{X}^{*}=\left\{ X\right\} $, an admissible super-mediator node is $\boldsymbol{M}^{*}=\left\{ M\right\} $ and an admissible super-effect node is $\boldsymbol{Y}^{*}=\left\{ Y\right\} $; in (i) and (j), an admissible super-cause node is $\boldsymbol{X}^{*}=\left\{ X\right\} $ and an admissible super-effect node is $\boldsymbol{Y}^{*}=\left\{ Y\right\} $, but the corresponding admissible super-mediator node $\boldsymbol{M}^{*}$ must be enlarged to $\{V, M \}$ and $\{U, M \}$, respectively; in (k), an admissible super-cause node needs enlargement to $\boldsymbol{X}^{*}=\left\{ X,U\right\} $, and the corresponding admissible super-mediator node is $\boldsymbol{M}^{*}=\left\{ M,V\right\} $ and an admissible super-effect node is $\boldsymbol{Y}^{*}=\left\{ Y\right\} $.
    \label{fig:FDR examples}}
\end{figure}

In Figure~\ref{fig:FDR examples}, we demonstrate some example ADMGs, where all of them can be reduced to an ordinary front-door setting with a proper construction (projection) of the super-cause, effect and mediator node. For instance, for Figure~\ref{fig:FDR examples} (f), we can prove that the interventional distribution $p\left(Y\left|do\left(X\right)\right.\right)$ has the same expression as the front-door adjustment formula in eq. \eqref{eq: ordinary fd adj formula} by considering $M$ as the mediator variable with the fundamental \emph{do}-calculus rules (eq. \eqref{eq: interventional dist derivation Figure 2 (f)} in Appendix~\ref{sec: interventional dist derivation Figure 2 (f)}).

On the other hand, if we test the front-door criterion on the ADMG in Figure~\ref{fig:FDR examples} (f) for the cause variable $X$, effect variable $Y$ and mediator variable $M$, all conditions are satisfied. This example is not ad hoc: a wide family of ADMGs, such as those in Figure~\ref{fig:FDR examples}, can be handled by front-door adjustment instead of applying the more complex ID algorithm and obtaining impractical expressions.

The question that follows is how do we test if a complicated ADMG can be reduced to the ordinary front-door setting? If it is reducible, how can we identify or construct admissible super-cause, effect and mediator node?

\subsection{Front-door reducibility (FDR)}

First, we formally define the \emph{front-door reducibility} as follows:

\begin{defn}[Front-door reducibility criterion, FDR criterion]
\label{def: front-door reducible}
An ADMG $\mathcal{G}\left(\boldsymbol{V}\right)$ relative to cause and effect variables $X$ and $Y$ is said to be front-door reducible to $\mathcal{G}_{\boldsymbol{X}^{*},\boldsymbol{Y}^{*}}^{*}$ , iff there exists a triple of disjoint and non-empty supersets $\left(\boldsymbol{X}^{*},\boldsymbol{Y}^{*},\boldsymbol{M}^{*}\right)$, referring to the super-cause, effect and mediator nodes, respectively, such that:

    \begin{itemize}
        \item \textbf{FDR1:} For any $X_{i}\in\boldsymbol{X}^{*}$ and any $Y_{i}\in\boldsymbol{Y}^{*}$, all variables on every directed path from $X_{i}$ to $Y_{i}$ intersect with $\boldsymbol{M}^{*}$, which is to say: $\forall X_{i}\in\boldsymbol{X}^{*}$, $\forall Y_{i}\in\boldsymbol{Y}^{*}$, $\forall\pi:X_{i}\rightarrow...\rightarrow Y_{i}$, $\pi\cap\boldsymbol{M}^{*}\neq\emptyset$, equivalently $\boldsymbol{Y}^{*}\cap\mathrm{De}_{\mathcal{G}\left(\boldsymbol{V}\setminus\boldsymbol{M}^*\right)}\left(\boldsymbol{X}^{*}\right)=\emptyset$;
        
        \item \textbf{FDR2:} For any $M_{i}\in\boldsymbol{M}^{*}$, there is no backdoor path between $\boldsymbol{X}^{*}$ and $M_{i}$, equivalently $\left(\boldsymbol{X}^{*}\perp_{m}\boldsymbol{M}^{*}\right)_{\mathcal{G}_{\underline{\boldsymbol{X}^{*}}}}$;
        
        \item \textbf{FDR3:} For any $M_{i}\in\boldsymbol{M}^{*}$, all backdoor paths between $\boldsymbol{Y}^{*}$ and $\boldsymbol{M}^{*}$ are blocked by $\boldsymbol{X}^{*}\cup\boldsymbol{M}_{-i}^{*}$, equivalently $\forall M_{i}\in\boldsymbol{M}^{*}$, $\left(\boldsymbol{Y}^{*}\perp_{m}M_{i}\left|\boldsymbol{X}^{*}\cup\boldsymbol{M}_{-i}^{*}\right.\right)_{\mathcal{G}_{\overline{\boldsymbol{X}^{*}}\underline{M_{i}}}}$.
    \end{itemize}
\end{defn}
The FDR criterion provides a shortcut so that we could use front-door adjustment with trivial substitution for variables to obtain the interventional distribution of interest. The following theorem proves that:
\begin{thm} 
    \label{theorem: FDR adjustment}\textbf{ (Front-door reducible (FDR) adjustment).} 
    If an ADMG $\mathcal{G}$ relative to cause and effect variables $X$ and $Y$ is front-door reducible , the causal effect of $\boldsymbol{X}^{*}$ on $\boldsymbol{Y}^{*}$ is identifiable by adjusting on $\boldsymbol{M}^{*}$. The interventional distribution $p\left(\boldsymbol{y}^{*}\left|do\left(\boldsymbol{x}^{*}\right)\right.\right)$ is given by the following FDR adjustment formula:
    
    \begin{equation}
        p\left(\boldsymbol{y}^{*}\left|do\left(\boldsymbol{x}^{*}\right)\right.\right)=\int p\left(\boldsymbol{y}^{*}\left|\boldsymbol{x'}^{*},\boldsymbol{m}^{*}\right.\right)p\left(\boldsymbol{m}^{*}\left|\boldsymbol{x}^{*}\right.\right)p\left(\boldsymbol{x'}^{*}\right)d\boldsymbol{m}^{*}d\boldsymbol{x'}^{*},
        \label{eq:fdr adj formula}
    \end{equation}
    where $\boldsymbol{X}^{*},\boldsymbol{Y}^{*},\boldsymbol{M}^{*}$ are super-cause, effect and mediator nodes, respectively.

    \begin{proof}
        Since the FDR criterion (FDR1–3) are exactly the standard front-door criterion applied to the super-nodes $\boldsymbol{X}^{*},\boldsymbol{Y}^{*},\boldsymbol{M}^{*}$, the reduced ADMG $\mathcal{G}_{\boldsymbol{X}^{*},\boldsymbol{Y}^{*}}^{*}$ has the FDR adjustment formula shown in eq. \eqref{eq:fdr adj formula} by simply substituting the variables with the super-node sets in eq. \eqref{eq: ordinary fd adj formula}.
        
        See Appendix~\ref{sec: FDR adjustment} for the detailed derivation for the FDR adjustment formula using three rules of \emph{do}-calculus.
    \end{proof}
\end{thm}
Theorem~\ref{theorem: FDR adjustment} shows that, for all possible ADMGs that satisfy FDR criterion, FDR adjustment formula exists, and its expression is given by eq. \eqref{eq:fdr adj formula}. Next, we prove that if there exists an interventional distribution in the form of FDR adjustment formula, the corresponding ADMG must satisfy FDR criterion. In other words, satisfaction of FDR criterion is equivalent to applicability FDR adjustment.  

\begin{thm}[Equivalence between FDR adjustment and FDR criterion satisfaction]
    \label{theorem: equivalence}
    Let $\mathcal{G}\left(\boldsymbol{V}\right)$ be an ADMG, $X$, $Y$ be the cause and effect variables of interest. The following statements are equivalent:

    \begin{enumerate}
        \item \textbf{FDR adjustment}: There exists a triple of disjoint and non-empty supersets $\left(\boldsymbol{X}^{*},\boldsymbol{Y}^{*},\boldsymbol{M}^{*}\right)$, such that the interventional distribution $p\left(\boldsymbol{y}^{*}\left|do\left(\boldsymbol{x}^{*}\right)\right.\right)$ equals the FDR adjustment formula eq. \eqref{eq:fdr adj formula} in $\mathcal{G}$.
        
        \item \textbf{FDR criterion}: There exists a triple of disjoint and non-empty supersets $\left(\boldsymbol{X}^{*},\boldsymbol{Y}^{*},\boldsymbol{M}^{*}\right)$, such that $\mathcal{G}\left(\boldsymbol{V}\right)$ is front-door reducible to $\mathcal{G}_{\boldsymbol{X}^{*},\boldsymbol{Y}^{*}}^{*}$.
    \end{enumerate}

    \begin{proof}
        See Appendix~\ref{sec: FDR criterion and adjustment equivalence}.
    \end{proof}
    
\end{thm}

Note that Theorem~\ref{theorem: equivalence} does not discuss ad-hoc numeric counterexamples, while it rests entirely on the \emph{do}-calculus rules. Consequently, if any such precondition fails, there exists a model compatible with $\mathcal{G}$ under which that equality breaks; hence eq. \eqref{eq:fdr adj formula} cannot serve as a universally valid formula in $\mathcal{G}$. More importantly, observing eq. \eqref{eq:fdr adj formula} which matches $p\left(\boldsymbol{y}^{*}\left|do\left(\boldsymbol{x}^{*}\right)\right.\right)$ for a single empirical distribution does not imply that $\mathcal{G}$ relative to cause and effect variables $X$ and $Y$ is front-door reducible to $\mathcal{G}_{\boldsymbol{X}^{*},\boldsymbol{Y}^{*}}^{*}$. The equivalence we establish is graph-level, which means eq. \eqref{eq:fdr adj formula} is required to hold for all semi-Markovian causal models with respect to $\mathcal{G}$, not merely for a specific dataset or parameterization.

\subsection{FDR triple}

Building on the FDR criterion and its equivalence to the FDR adjustment, we now characterize how to algorithmically construct admissible super-nodes (which we refer to as the \emph{FDR triple}) $\left(\boldsymbol{X}^{*},\boldsymbol{Y}^{*},\boldsymbol{M}^{*}\right)$. To this end, we define a search space of candidate triples (Definition~\ref{def: FDR triple}), and then show its correctness (Theorem~\ref{Theorem: FDR triple correctness}) and completeness (Lemma~\ref{lemma:Completeness of X}-\ref{lemma: Completeness of Z_S} and Theorem~\ref{theorem: FDR triple completeness}). The following definition gives a formal construction for the \emph{FDR triple} mentioned above, searching through a constructive space whose admissible sets are exactly the FDR triples satisfying the FDR criterion.

\begin{defn}[FDR triple]
    \label{def: FDR triple} 
    Let $\mathcal{G}\left(\boldsymbol{V}\right)$ be an ADMG, $X,Y\in\boldsymbol{V}$ be the cause and effect of interest. The super-effect node $\boldsymbol{Y}^{*}$ is simply:
    
    \begin{equation}
    \boldsymbol{Y}^{*}=\left\{ Y\right\} 
    \end{equation}
    
    Define the potential super-cause family $\boldsymbol{\mathcal{X}}$ as a set of candidate super-cause set by:
    
    \begin{equation}
        \boldsymbol{\mathcal{X}}=\left\{ \boldsymbol{S}\ensuremath{\subseteq\left\{ X\right\} \cup\left(\mathrm{An}_{\mathcal{G}}\left(\boldsymbol{Y}^{*}\right)\setminus\boldsymbol{Y}^{*}\right)}:X\in\boldsymbol{S}\right\} 
        \label{eq:super-cause family}
    \end{equation}
    
    For $\boldsymbol{S}'\in\boldsymbol{\mathcal{X}}$, define the candidate mediator region relative to $\boldsymbol{S}'$ as $\boldsymbol{Z}\left(\boldsymbol{S}'\right)$ by:
    
    \begin{equation}
        \boldsymbol{Z}\left(\boldsymbol{S}'\right)=\left(\mathrm{An}_{\mathcal{G}}\left(\boldsymbol{Y}^{*}\right)\cap\mathrm{De}_{\mathcal{G}}\left(\boldsymbol{S}'\right)\right)\setminus\left(\mathrm{Nb}_{\mathcal{G}}^{\leftrightarrow}\left(\boldsymbol{S}'\right)\cup\mathrm{Nb}_{\mathcal{G}}^{\leftrightarrow}\left(\boldsymbol{Y}^{*}\right)\cup\boldsymbol{S}'\cup\boldsymbol{Y}^{*}\right)
        \label{eq:candidate mediator region}
    \end{equation}
    
    define the candidate super-mediator family relative to $\boldsymbol{S}'$ as $\boldsymbol{\mathcal{M}}\left(\boldsymbol{S}'\right)$ by:
    
    \begin{align}
        \boldsymbol{\mathcal{M}}\left(\boldsymbol{S}'\right)= & \left\{ \boldsymbol{M}\subseteq\boldsymbol{Z}\left(\boldsymbol{S}'\right):\underbrace{\boldsymbol{Y}^{*}\cap\mathrm{De}_{\mathcal{G}\left(\boldsymbol{V}\setminus\boldsymbol{M}\right)}\left(\boldsymbol{S}'\right)=\emptyset}_{\text{FDR 1}}\land\underbrace{\left(\boldsymbol{S}'\perp_{m}\boldsymbol{M}\right)_{\mathcal{G}_{\underline{\boldsymbol{S}'}}}}_{\text{FDR 2}}\land\right.\nonumber \\
         & \left.\underbrace{\forall M_{i}\in\boldsymbol{M}:\left(\boldsymbol{Y}^{*}\perp_{m}M_{i}\left|\boldsymbol{S}'\cup\boldsymbol{M}_{-i}\right.\right)_{\mathcal{G}_{\overline{\boldsymbol{S}'}\underline{M_{i}}}}}_{\text{FDR 3}}\right\} 
         \label{eq:candidate super-mediator family}
    \end{align}
    
    If there exists a $\boldsymbol{S}'$ such that it induces a non-empty set $\boldsymbol{M}^{'}\in\boldsymbol{\mathcal{M}}\left(\boldsymbol{S}'\right)$, then we define the super-cause node $\boldsymbol{X}^{*}$ and the super-mediator node as $\boldsymbol{M}^{*}$ by:
    
    \begin{equation}
        \boldsymbol{X}^{*}=  \boldsymbol{S}', 
        \quad
        \boldsymbol{M}^{*}=  \boldsymbol{M}^{'}
    \end{equation}
    
    We call $\left(\boldsymbol{X}^{*},\boldsymbol{Y}^{*},\boldsymbol{M}^{*}\right)$ as the FDR triple of $\mathcal{G}$ relative to $\left(X,Y\right)$. Meanwhile, by construction, $\boldsymbol{X}^{*},\boldsymbol{Y}^{*},\boldsymbol{M}^{*}$ are pairwise disjoint.
\end{defn}

In the above construction, we fix $\boldsymbol{Y}^{*}=\left\{ Y\right\} $ without any enlargement, though intuitively in some scenarios we may consider enlarging the super-effect node. For example, in Figure~\ref{fig:FDR examples} (k), due to the bidirected edge connecting $Z$ and $Y$, we may consider an enlarged super-effect node $\boldsymbol{Y}^{+}=\left\{ Z,Y\right\} $ to reduce the ADMG with $\boldsymbol{M}^{*}=\left\{ M,V\right\} $ and $\boldsymbol{X}^{*}=\left\{ X,U\right\} $. In fact, the triple $\left(\boldsymbol{X}^{*}=\left\{ X,U\right\} ,\boldsymbol{Y}^{+}=\left\{ Z,Y\right\} ,\boldsymbol{M}^{*}=\left\{ M,V\right\} \right)$ also satisfies the FDR criterion in Definition~\ref{def: front-door reducible}. Nevertheless, in this example, the triple $\left(\left\{ X,U\right\} ,\left\{ Y\right\} ,\left\{ M,V\right\} \right)$ is still sufficient to satisfy FDR criterion, where (i) $\boldsymbol{M}^{*}$ intercepts all directed paths from $\boldsymbol{X}^{*}$ to $Y$; (ii) there is no backdoor between $\boldsymbol{X}^{*}$ and $\boldsymbol{M}^{*}$; (iii) backdoor paths between $Y$ and $M$ are blocked by conditioning on $\boldsymbol{X}^{*}\cup\left\{ V\right\} $ and backdoor paths between $Y$ and $V$ are blocked by conditioning on $\boldsymbol{X}^{*}\cup\left\{ M\right\} $. Without loss of generality, we have the proposition of effect minimality:

\begin{proposition}[Effect minimality]
    \label{Prop: effect mini}
    Let $\mathcal{G}\left(\boldsymbol{V}\right)$ be an ADMG and $\left(X,Y\right)$ be the cause-effect pair of interest. If there exists a super effect node $Y\in\boldsymbol{Y}^{+}$ and corresponding triple $\left(\boldsymbol{X}^{*},\boldsymbol{Y}^{+},\boldsymbol{M}^{*}\right)$ satisfies FDR criterion, then the triple $\left(\boldsymbol{X}^{*},\boldsymbol{Y}^{*}=\left\{ Y\right\} ,\boldsymbol{M}^{*}\right)$ also satisfies FDR criterion. Hence, w.l.o.g., we fix $\boldsymbol{Y}^{*}=\left\{ Y\right\} $.

    \begin{proof}
        See Appendix~\ref{sec: effect minimality}.
    \end{proof}
    
\end{proposition}

Next, we show the correctness of the construction of the FDR triple $\left(\boldsymbol{X}^{*},\boldsymbol{Y}^{*},\boldsymbol{M}^{*}\right)$ in Definition~\ref{def: FDR triple} must satisfy FDR criterion.

\begin{thm}[Correctness of the FDR triple construction]
    \label{Theorem: FDR triple correctness}
    If there exists a triple of non-empty supersets $\left(\boldsymbol{X}^{*},\boldsymbol{Y}^{*},\boldsymbol{M}^{*}\right)$ constructed as Definition~\ref{def: FDR triple} for an ADMG $\mathcal{G}\left(\boldsymbol{V}\right)$ with cause and effect of interest $X$ and $Y$, then $\left(\boldsymbol{X}^{*},\boldsymbol{Y}^{*},\boldsymbol{M}^{*}\right)$ must satisfy the FDR criterion.

    \begin{proof}
        The construction of the candidate super-mediator family relative to $\boldsymbol{S}'$ as $\boldsymbol{\mathcal{M}}\left(\boldsymbol{S}'\right)$ in Definition~\ref{def: FDR triple} is formulated with the same conditions as FDR 1-3 in Definition~\ref{def: front-door reducible}. Therefore, given an ADMG $\mathcal{G}\left(\boldsymbol{V}\right)$, if there exists a FDR triple of non-empty supersets $\left(\boldsymbol{X}^{*},\boldsymbol{Y}^{*},\boldsymbol{M}^{*}\right)$ constructed as shown in Definition~\ref{def: FDR triple}, then the ADMG $\mathcal{G}\left(\boldsymbol{V}\right)$ relative to cause and effect variables of interest $X$ and $Y$ is said to be front-door reducible to $\mathcal{G}_{\boldsymbol{X}^{*},\boldsymbol{Y}^{*}}^{*}$.
    \end{proof}
    
\end{thm}
With the FDR triple correctness, we now have an important corollary to show the applicability of FDR adjustment shown in eq. \eqref{eq:fdr adj formula} to an ADMG which is FDR.

\begin{cor}[Applicability of FDR adjustment on the FDR triple]
    Given an ADMG $\mathcal{G}\left(\boldsymbol{V}\right)$ with the cause and effect of interest $X$ and $Y$, if there exist a triple of non-empty supersets $\left(\boldsymbol{X}^{*},\boldsymbol{Y}^{*},\boldsymbol{M}^{*}\right)$ constructed as Definition~\ref{def: FDR triple}, the interventional distribution $p\left(\boldsymbol{Y}^{*}\left|do\left(\boldsymbol{X}^{*}\right)\right.\right)$ is given by eq. \eqref{eq:fdr adj formula}.

    \begin{proof}
        The construction of the FDR triple shown in Definition~\ref{def: FDR triple} is based on the FDR criterion, i.e., the three conditions are explicitly included when constructing the candidate super-mediator family $\boldsymbol{\mathcal{M}}\left(\boldsymbol{S}'\right)$.
    \end{proof}
\end{cor}
Meanwhile, the construction of FDR triple shown in Definition~\ref{def: FDR triple} also includes all possible triples that makes an ADMG front-door reducible. Before we provide the completeness of such a construction, we show the completeness for the potential super-cause family $\boldsymbol{\mathcal{X}}$ and the candidate mediator region $\boldsymbol{Z}\left(\boldsymbol{S}'\right)$ first. 

\begin{lem}[Completeness of the potential super-cause family $\boldsymbol{\mathcal{X}}$]
    \label{lemma:Completeness of X}
    Given an ADMG $\mathcal{G}\left(\boldsymbol{V}\right)$, suppose there exists a triple $\left(\boldsymbol{X}^{+},\boldsymbol{Y}^{*},\boldsymbol{M}^{+}\right)$ satisfying FDR criterion. Then there exists:
    
    \begin{equation}
        \boldsymbol{X}'=\left\{ X\right\} \cup\left(\boldsymbol{X}^{+}\cap\left(\mathrm{An}\left(\boldsymbol{Y}^{*}\right)\setminus\boldsymbol{Y}^{*}\right)\right)
    \end{equation}
    
    Then, $\boldsymbol{X}'\in\boldsymbol{\mathcal{X}}$, and there exists some $\boldsymbol{M}'\subseteq\boldsymbol{M}^{+}$ such that the triple $\left(\boldsymbol{X}^{'},\boldsymbol{Y}^{*},\boldsymbol{M}^{'}\right)$ is a FDR triple. That is to say, the potential super-cause family $\boldsymbol{\mathcal{X}}$ is \textbf{complete}.

    \begin{proof}
        See Appendix~\ref{sec: Completeness of the potential super-cause family}.
    \end{proof}

\end{lem}
Lemma~\ref{lemma:Completeness of X} comes with the conclusion that only the nodes that can be potentially merged into the super-cause node in a given ADMG $\mathcal{G}\left(\boldsymbol{V}\right)$ will be included in $\boldsymbol{\mathcal{X}}$. For example, in Figure~\ref{fig:FDR examples} (d), $\boldsymbol{\mathcal{X}}=\left\{ \left\{ X\right\} ,\left\{ X,M\right\} \right\} $. In other words, $\nexists\boldsymbol{S}\in\boldsymbol{\mathcal{X}}$ such that $Y\in\boldsymbol{S}$ or $U\in\boldsymbol{S}$ or $V\in\boldsymbol{S}$. Intuitively, we don't consider $V$ and $U$ potentially in the super-cause node because the path $\left(X\leftrightarrow V\leftarrow U\leftarrow Y\right)$ is naturally blocked by the collider $V$, and including any of $U$ or $V$ (or both) in the super-cause node will break the FDR criterion. 

\begin{lem}[Completeness of the candidate mediator region $\boldsymbol{Z}\left(\boldsymbol{S}'\right)$]
    \label{lemma: Completeness of Z_S}
    Given an ADMG $\mathcal{G}\left(\boldsymbol{V}\right)$, let $\boldsymbol{S}'\in\boldsymbol{\mathcal{X}}$ as Definition~\ref{def: FDR triple}. Suppose there exists $\boldsymbol{M}^{+}\subseteq\boldsymbol{V}$ such that the triple $\left(\boldsymbol{S}',\boldsymbol{Y}^{*},\boldsymbol{M}^{+}\right)$ satisfies FDR criterion. Then there exists:
    
    \begin{equation}
        \boldsymbol{M}'\subseteq\boldsymbol{Z}\left(\boldsymbol{S}'\right)=\left(\mathrm{An}_{\mathcal{G}}\left(\boldsymbol{Y}^{*}\right)\cap\mathrm{De}_{\mathcal{G}}\left(\boldsymbol{S}'\right)\right)\setminus\left(\mathcal{\mathrm{Nb}_{\mathcal{G}}^{\leftrightarrow}}\left(\boldsymbol{S}'\right)\cup\mathrm{Nb}_{\mathcal{G}}^{\leftrightarrow}\left(\boldsymbol{Y}^{*}\right)\cup\boldsymbol{S}'\cup\boldsymbol{Y}^{*}\right),
    \end{equation}
    such that an FDR triple $\left(\boldsymbol{S}',\boldsymbol{Y}^{*},\boldsymbol{M}'\right)$ also satisfies the FDR criterion. That is to say, the candidate mediator region $\boldsymbol{Z}\left(\boldsymbol{S}'\right)$ is \textbf{complete}.

    \begin{proof}
        See Appendix~\ref{sec: Completenesss of the candidate mediator region}.
    \end{proof}
    
\end{lem}
Lemma~\ref{lemma: Completeness of Z_S} comes with the conclusion that only the nodes that can be potentially merged into the super-mediator node in a given ADMG $\mathcal{G}\left(\boldsymbol{V}\right)$ will be included in $\boldsymbol{Z}\left(\boldsymbol{S}'\right)$. For example, in Figure~\ref{fig:FDR examples} (e), given $\boldsymbol{S}'=\left\{ X\right\} $, the candidate mediator region relative to $\boldsymbol{S}'$ is $\boldsymbol{Z}\left(\boldsymbol{S}'\right)=\left\{ M\right\} $. In other words, $V\notin\boldsymbol{Z}\left(\boldsymbol{S}'\right)$ or $U\notin\boldsymbol{Z}\left(\boldsymbol{S}'\right)$. Intuitively, adjusting on a super-mediator node with any (or both) of them open the backdoor path between $Y$ and the actual mediator $M$ when conditioning on $X$ in $\mathcal{G}_{\overline{X}\underline{M}}$, and also open the backdoor path between $X$ and $M$ in $\mathcal{G}_{\underline{X}}$. 

\begin{thm}[Completeness of FDR triple for FDR criterion]
    \label{theorem: FDR triple completeness}
    Given an ADMG $\mathcal{G}\left(\boldsymbol{V}\right)$ with at least one directed path $\pi_{X\rightarrow Y}$ and the cause and effect of interest $X,Y\in V$, if there does \textbf{not} exist a FDR triple of non-empty sets $\left(\boldsymbol{X}^{*},\boldsymbol{Y}^{*},\boldsymbol{M}^{*}\right)$ constructed as stated in the Definition~\ref{def: FDR triple}, such that:
    
    \begin{align}
        \emptyset\neq\boldsymbol{X}^{*} & \in\boldsymbol{\mathcal{X}}\text{, and}\nonumber \\
        \boldsymbol{Y}^{*} & =\left\{ Y\right\} \text{, and}\\
        \emptyset\neq\boldsymbol{M}^{*} & \subseteq\boldsymbol{Z}\left(\boldsymbol{X}^{*}\right),\nonumber 
    \end{align}
    then the $\mathcal{G}\left(\boldsymbol{V}\right)$ is \textbf{not} front-door reducible. Equivalently, the construction of Definition~\ref{def: FDR triple} is complete for the FDR criterion.

    \begin{proof}
        By Lemma~\ref{lemma:Completeness of X}, we obtain $\boldsymbol{X}^{*}\in\boldsymbol{\mathcal{X}}$ with necessary and harmless shrinking. By Proposition~\ref{Prop: effect mini}, we proved the minimal super-effect is $\boldsymbol{Y}^{*}=\left\{ Y\right\} $. By Lemma~\ref{lemma: Completeness of Z_S}, we can prune mediators to some $\boldsymbol{M}^{*}\subseteq\boldsymbol{Z}\left(\boldsymbol{X}^{*}\right)$ while preserving FDR criterion.
    \end{proof}
    
\end{thm}
\begin{cor}[The irrelevance of FDR for strict descendants of $\boldsymbol{Y}^*$]
    \label{corol: irrelevance for De(Y)} 
    Let $\mathcal{G}\left(\boldsymbol{V}\right)$ be an ADMG with cause and effect variables of interest $X,Y$. Let $\mathcal{G}^{-}$ be the induced subgraph on $\left\{ X\right\} \cup\left(\mathrm{An}_{\mathcal{G}}\left(\boldsymbol{Y}^{*}\right)\right)$ . We use the term \textbf{strict descendants} of $\boldsymbol{Y}^{*}$ here to refer $V_{i}\in\mathrm{De}_{\mathcal{G}}\left(\boldsymbol{Y}^{*}\right)\setminus\boldsymbol{Y}^{*}$. Then:
    
    \begin{enumerate}
        \item $\mathcal{G}$ relative to cause and effect variables $X$ and $Y$ is front-door reducible iff. $\mathcal{G}^{-}$ relative to cause and effect variables $X$ and $Y$ is front-door reducible.
        
        \item All admissible FDR triples as constructed in Definition~\ref{def: FDR triple} $\left(\boldsymbol{X}^{*},\boldsymbol{Y}^{*},\boldsymbol{M}^{*}\right)$ in $\mathcal{G}$ are also all admissible FDR triples in $\mathcal{G}^{-}$.
    \end{enumerate}
    
    In other words, \textbf{$\mathrm{De}_{\mathcal{G}}\left(Y\right)$} is irrelevant to whether an ADMG satisfies FDR or not, and deleting or adding descendant nodes of $Y$ does not change the admissible set of FDR triples.

    \begin{proof}
            By Definition~\ref{def: FDR triple}, the potential super-cause family $\boldsymbol{\mathcal{X}}$ only contains the region $\boldsymbol{R}=\left\{ X\right\} \cup\left(\mathrm{An}_{\mathcal{G}}\left(\boldsymbol{Y}^{*}\right)\setminus\boldsymbol{Y}^{*}\right)$ as shown in eq. \eqref{eq:super-cause family}, and the candidate mediator region $\boldsymbol{Z}$ also contains the region at most $\mathrm{An}_{\mathcal{G}}\left(\boldsymbol{Y}^{*}\right)$. That means the construction of an admissible FDR triple is solely on the region $\boldsymbol{R}\cup\boldsymbol{Y}^{*}$. Meanwhile, the FDR criterion stated as Definition~\ref{def: front-door reducible} shows that all m-separation tests are for nodes in the region $\boldsymbol{R}\cup\boldsymbol{Y}^{*}$. If there exists a strict descendant $V_{i}$ such that attempts to traverse to the region $\boldsymbol{R}$ from $\boldsymbol{Y}^{*}$, it either breaks the fundamental acyclicity for ADMG or hits a collider on the path, which is naturally blocked without conditioning on the collider or its strict descendants. That proves $\mathcal{G}$ and $\mathcal{G}^{-}$ are equivalent for FDR criterion satisfaction and FDR triple existence.
    \end{proof}
    
\end{cor}
Corollary~\ref{corol: irrelevance for De(Y)} is very useful to simplify the ADMG, for example, the descendants of $Y$ in Figure~\ref{fig:FDR examples} (b), (d) and (e) can be removed harmlessly for FDR criterion test and FDR triple construction. In Figure~\ref{fig:FDR examples} (b), $V$ is the only strict descendant of $Y$, and it is obvious that the path $\left(Y\rightarrow V\leftrightarrow M\right)$ is blocked by the collider $V$ itself. Removing it from the ADMG does not affect FDR criterion test and also the construction of an admissible FDR triple, i.e., the only admissible FDR triple is given by $\left(\boldsymbol{X}^{*}=\left\{ X\right\}, \boldsymbol{Y}^{*}=\left\{ Y\right\}, \boldsymbol{M}^{*}=\left\{ M\right\} \right)$.

\subsection{Non-FDR examples}

\begin{figure}[htbp]
    \begin{centering}
    \includegraphics[width=0.7\linewidth]{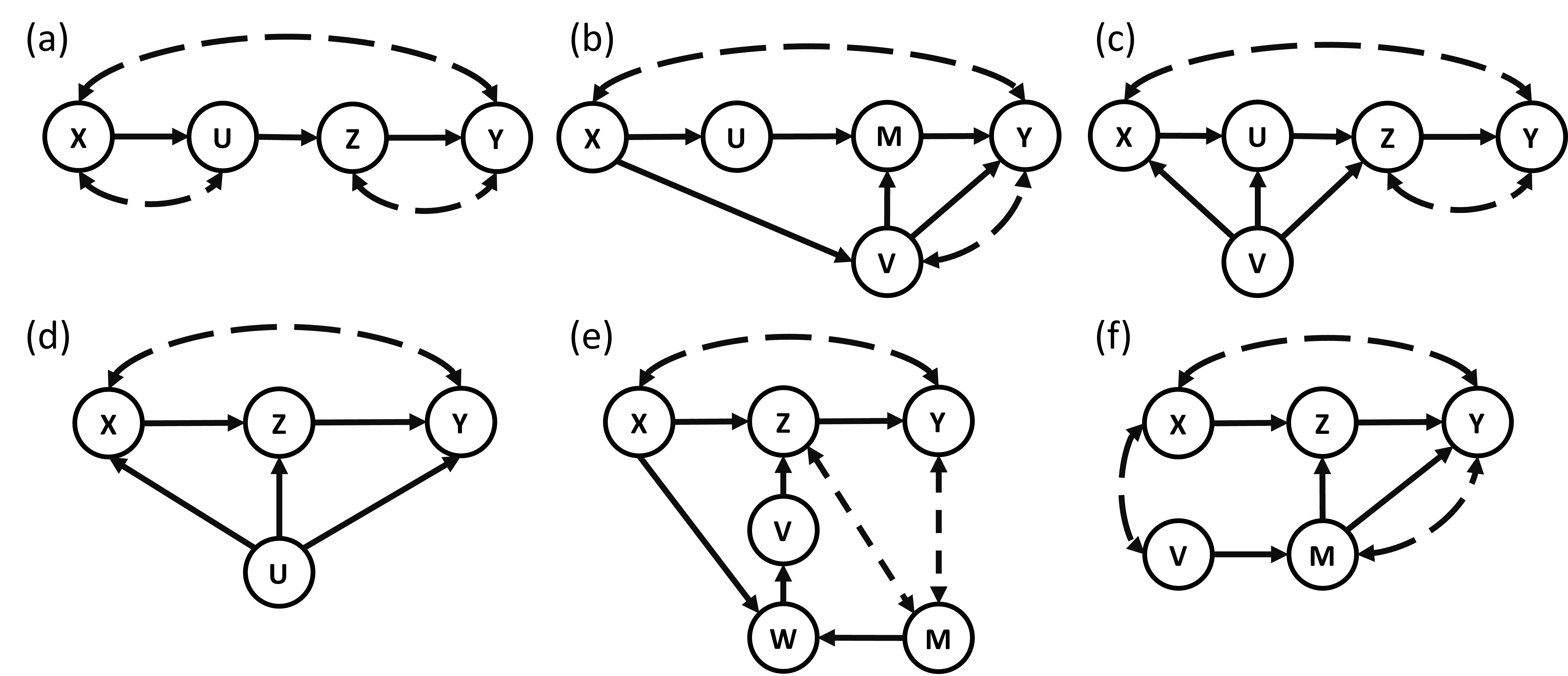}
    \par\end{centering}
    \caption{Example ADMGs relative to the cause and effect of interest $X,Y$
    not satisfying FDR criterion. 
    \label{fig:non-FDR examples}}
\end{figure}

Figure~\ref{fig:non-FDR examples} demonstrates some examples that do not satisfy the FDR criterion. Let's check one of the failure example as shown in Figure~\ref{fig:non-FDR examples} (b). Using the construction in Definition~\ref{def: FDR triple}, the super-effect node $\boldsymbol{Y}^{*}=\left\{ Y\right\} $. In the potential super-cause family $\boldsymbol{\mathcal{X}}$ has 8 elements (sets), it is easy to figure out none of them $\boldsymbol{S}'\in\boldsymbol{\mathcal{X}}=\left\{ \boldsymbol{S}\subseteq\left\{ X,U,M,V\right\} :X\in\boldsymbol{S}\right\} $ can lead to a non-empty $\boldsymbol{M}^{*}\in\boldsymbol{\mathcal{M}}\left(\boldsymbol{S}'\right)$. Let's explore the two candidate super-cause nodes that seems most likely to satisfy FDR, i.e., (i) $\boldsymbol{S}'_{1}=\left\{ X\right\} $; and (ii) $\boldsymbol{S}'_{2}=\left\{ X,V\right\} $:

\begin{enumerate}
    \item $\boldsymbol{S}'_{1}=\left\{ X\right\} $: Using eq. \eqref{eq:candidate mediator region}, we have $\boldsymbol{Z}\left(\boldsymbol{S}'_{1}\right)=\left\{ U,M\right\} $. Then, it is obvious that adjusting either solely on $U$, $M$ or on their joint, there exists an directed path $\left(X\rightarrow V\rightarrow Y\right)$. This breaks the FDR 1 in eq. \eqref{eq:candidate super-mediator family}, which it leads to a candidate super-mediator family with only an empty set, i.e., $\boldsymbol{\mathcal{M}}\left(\boldsymbol{S}'_{1}\right)=\left\{ \emptyset\right\} $. 
    
    \item $\boldsymbol{S}'_{2}=\left\{ X,V\right\} $: Using eq. \eqref{eq:candidate mediator region}, we have $\boldsymbol{Z}\left(\boldsymbol{S}'_{2}\right)=\left\{ M\right\} $. There is a directed path $V\rightarrow Y$ which cannot be intercepted by $M$. That also breaks the FDR 1, so $\boldsymbol{\mathcal{M}}\left(\boldsymbol{S}'_{2}\right)=\left\{ \emptyset\right\} $.
\end{enumerate}
So there is no an admissible super-mediator set $\boldsymbol{M}^{*}$ for the FDR criterion in the above setting. However, non-FDR does not imply a causal relation of interest is not identifiable. For example, $p\left(y\left|do\left(x\right)\right.\right)$ is still identifiable using the ID algorithm \citep{shpitser2006identification} for Figure~\ref{fig:non-FDR examples} (c), (d) and (f), but they are non-FDR. As a simple example, the interventional distribution $p\left(y\left|do\left(x\right)\right.\right)$ for the ADMG shown in Figure~\ref{fig:non-FDR examples} (c) can be written as:

\begin{equation}
    p\left(y\left|do\left(x\right)\right.\right)=\int p\left(y\left|u,v,x'\right.\right)p\left(u\left|v,x\right.\right)p\left(v,x'\right)dudvdx',
\end{equation}
which cannot be reduced to an FDR adjustment formula by either trivial simplification or by grouping a few variables as a super-node.

\section{Admissible FDR triple identification algorithm}

Given the above discussion and formal definition along with proved properties of FDR and FDR triple, we now attempt to solve the FDR triple identification problem if an ADMG is FDR. Based on aforementioned properties and FDR triple definition, we propose an admissible FDR triple identification (FDR-TID) algorithm as shown in Algorithm~\ref{alg: Admissible-FDR-triple}.    

\begin{algorithm}[ht]
\caption{Admissible FDR triple identification (FDR-TID)}
\label{alg: Admissible-FDR-triple}
    \begin{algorithmic}[1] 
    \item[] \textbf{Input:} An ADMG $\mathcal{G}\left(\boldsymbol{V} \right)$, the cause and effect variable pair of interest $\left(X, Y \right)$.         
    \item[] \textbf{Output:} FDR triple $ \left(\boldsymbol{X}^{*},\boldsymbol{Y}^{*},\boldsymbol{M}^{*}\right)$ or FAIL
    
    \State Fix: $\boldsymbol{Y}^{*}=\left\{ Y\right\}$ \Comment{Proposition 1}     
                
    \State Shrink to subgraph: $\mathcal{G}^{-} = \mathcal{G}\left(\left\{ X\right\} \cup\mathrm{An}_{\mathcal{G}}\left(\boldsymbol{Y}^{*}\right)\right)$ \Comment{Corollary 2}
    
    \For{$\boldsymbol{S'}\subseteq\left\{ X\right\} \cup\left(\mathrm{An}_{\mathcal{G}^{-}}\left(\boldsymbol{Y}^{*}\right)\setminus\boldsymbol{Y}^{*}\right)$ s.t. $X \in \boldsymbol{S'}$} \Comment{Lemma 1}
                         
    	\State Identify interior nodes: $\boldsymbol{I} = {\left(\mathrm{An}_{\mathcal{G}^{-}}\left(\boldsymbol{Y}^{*}\right)\cap\mathrm{De}_{\mathcal{G}^{-}}\left(\boldsymbol{S}'\right)\right)}\setminus\left(\boldsymbol{S}'\cup\boldsymbol{Y}^{*}\right)$                          
    
        \State Identify the candidate mediator region relative to $\boldsymbol{S}'$:             
    	\Statex \hfill $\boldsymbol{Z}\left(\boldsymbol{S}'\right) = \boldsymbol{I} \setminus\left(\mathrm{Nb}_{\mathcal{G}^{-}}^{\leftrightarrow}\left(\boldsymbol{S}'\right)\cup\mathrm{Nb}_{\mathcal{G}^{-}}^{\leftrightarrow}\left(\boldsymbol{Y}^{*}\right)\right)$  \hfill~ \Comment{Lemma 2}
    
    	\For{$\boldsymbol{M'}\subseteq\boldsymbol{Z}\left(\boldsymbol{S}'\right)$}
    		\If{ $\boldsymbol{Y}^{*}\cap\mathrm{De}_{\mathcal{G}^{-}\left(\boldsymbol{V}_{\mathcal{G}^{-}}\setminus\boldsymbol{M'}\right)}\left(\boldsymbol{S}'\right)\neq\emptyset$}                 
    			\State \textbf{continue} \Comment{If FDR 1 fails to hold, skip to the next $M'$ search.}      
    		\EndIf
    
            \If{$\left(\boldsymbol{S}' \not\perp_{m}\boldsymbol{M'}\right)_{\mathcal{G}^{-}_{\underline{\boldsymbol{S}'}}}$}                     
    			\State \textbf{continue} \Comment{If FDR 2 fails to hold, skip to the next $M'$ search.}                 
    		\EndIf
    
    		\If{$\exists M_{i}\in\boldsymbol{M'}:\left(\boldsymbol{Y}^{*}\not\perp_{m}M_{i}\left|\boldsymbol{S}'\cup\boldsymbol{M}_{-i}\right.\right)_{\mathcal{G}^{-}_{\overline{\boldsymbol{S}'}\underline{M_{i}}}}$}                     
    			\State \textbf{continue} \Comment{If FDR 3 fails to hold, skip to the next $M'$ search.}                 
    		\EndIf
                    
    		\Return $\left(\boldsymbol{X}^{*}=\boldsymbol{S}',  \boldsymbol{Y}^{*}=\left\{ Y\right\}, \boldsymbol{M}^{*}= \boldsymbol{M'} \right)$             
    	\EndFor         
    \EndFor      
      
    \Return FAIL   
         
    \end{algorithmic}
\end{algorithm}     

Before establishing the correctness and completeness of FDR-TID, we briefly illustrate how it operates on the ADMG $\mathcal{G}\left(\boldsymbol{V}\right)$ in Figure~\ref{fig:FDR examples} (j), where we want to identify a FDR triple for the cause and effect variable pair $\left(X,Y\right)$.

Line 1 fixes $\boldsymbol{Y}^{*}=\left\{ Y\right\} $, and $\mathcal{G}\left(\boldsymbol{V}\right)$ cannot be shrunk at line 2 since $\boldsymbol{V}=\left\{ X\right\} \cup\mathrm{An}_{\mathcal{G}}\left(\boldsymbol{Y}^{*}\right)$, hence $\mathcal{G}^{-}=\mathcal{G}$. When the algorithm invokes the outer loop for $\boldsymbol{S}'$ at line 3, let's start with $\boldsymbol{S}'=\left\{ X\right\} \subseteq\left\{ X\right\} \cup\mathrm{An}_{\mathcal{G}}\left(\boldsymbol{Y}^{*}\right)$. Line 5 gives the candidate mediator region relative to $\boldsymbol{S}'=\left\{ X\right\} $: $\boldsymbol{Z}_{\boldsymbol{S}'}=\left\{ U,M\right\} $. In the inner loop (line 6) for $\boldsymbol{M}'$, let's start from $\boldsymbol{M}'=\left\{ U\right\} \subseteq\boldsymbol{Z}_{\boldsymbol{S}'}$ without loss of generality. Here, $\mathrm{De}_{\mathcal{G}^{-}\left(\boldsymbol{V}_{\mathcal{G}^{-}}\setminus\left\{ U\right\} \right)}\left(\left\{ X\right\} \right)=\left\{ M,X,Y,V\right\} $, so that $\boldsymbol{Y}^{*}\cap\mathrm{De}_{\mathcal{G}^{-}\left(\boldsymbol{V}_{\mathcal{G}^{-}}\setminus\left\{ U\right\} \right)}\left(X\right)=\left\{ Y\right\} \neq\emptyset$. FDR 1 check does not pass, which means $U$ fails to intercepts all directed path from $\boldsymbol{S}'$ to $\boldsymbol{Y}^{*}$. Then the algorithm goes to the next iteration by selecting $\boldsymbol{M}'=\left\{ M\right\} \subseteq\boldsymbol{Z}_{\boldsymbol{S}'}$. In this case, $\mathrm{De}_{\mathcal{G}^{-}\left(\boldsymbol{V}_{\mathcal{G}^{-}}\setminus\left\{ M\right\} \right)}\left(X\right)=\left\{ X,Y,V,U\right\} $, and $\boldsymbol{Y}^{*}\cap\mathrm{De}_{\mathcal{G}^{-}\left(\boldsymbol{V}_{\mathcal{G}^{-}}\setminus\left\{ M\right\} \right)}\left(X\right)=\left\{ Y\right\} \neq\emptyset$, again FDR 1 check does not pass. Then the algorithm extends set $\boldsymbol{M}'=\left\{ U,M\right\} \subseteq\boldsymbol{Z}_{\boldsymbol{S}'}$. This leads to $\mathrm{De}_{\mathcal{G}^{-}\left(\boldsymbol{V}_{\mathcal{G}^{-}}\setminus\left\{ U,M\right\} \right)}\left(\left\{ X\right\} \right)=\left\{ X\right\} $, so $\boldsymbol{Y}^{*}\cap\mathrm{De}_{\mathcal{G}^{-}\left(\boldsymbol{V}_{\mathcal{G}^{-}}\setminus\left\{ U,M\right\} \right)}\left(X\right)=\emptyset$, passing FDR 1 check. After that, the algorithm goes to check FDR 2 at line 10. After deleting all outgoing arcs from $X$, it is clear that there is no backdoor paths between $\boldsymbol{S}'$ and $\boldsymbol{M}'$, such that $\left(X\perp_{m}U,M\right)_{\mathcal{G}_{\underline{X}}^{-}}$, passing FDR 2 check. Checking FDR 3 at line 13, for both $U$ and $M$, graphical preconditions $\left(Y\perp_{m}M\left|X,U\right.\right)_{\mathcal{G}_{\overline{X}\underline{M}}^{-}}$ and $\left(Y\perp_{m}U\left|X,M\right.\right)_{\mathcal{G}_{\overline{X}\underline{U}}^{-}}$ are satisfied. FDR 3 check passes. Finally, the FDR-TID returns an admissible FDR triple $\left(\left\{ X\right\} ,\left\{ Y\right\} ,\left\{ U,M\right\} \right)$.

It is worth mentioning that for other ADMGs, such as those shown in Figure \ref{fig:FDR examples} (a-c) and (g-i), can have multiple admissible FDR triples as illustrated in Table \ref{tab:All-admissible-FDR}. The FDR-TID algorithm returns the first admissible FDR triple when it is found, whereas this does not affect the correctness and completeness of the proposed FDR-TID. We next give formal discussion about the algorithm's correctness and completeness.

\begin{thm}[Correctness of FDR-TID]
\label{theorem: correctness of FDR-TID}
    If FDR-TID returns a triple $\left(\boldsymbol{X}^{*},\boldsymbol{Y}^{*},\boldsymbol{M}^{*}\right)$, then it must be a FDR triple as defined in Definition~\ref{def: FDR triple}.

    \begin{proof}
        FDR-TID follows the constructions in Definition~\ref{def: FDR triple} whose correctness has been proved in Theorem~\ref{Theorem: FDR triple correctness}. The algorithm enumerates every pair $\left(\boldsymbol{S}',\boldsymbol{M}'\right)$ (line 3 and 6), and checks the three FDR conditions in eq. \eqref{eq:candidate super-mediator family} (line 7-15), equivalent to the defined construction of the FDR triple.
    \end{proof}

\end{thm}
Next, we provide a lemma to show FDR-TID termination and analyse its time complexity.
\begin{lem}[FDR-TID finite search space and termination]
    \label{lemma: FDR-TID termination}
    Let $\mathcal{G}\left(\boldsymbol{V}\right)$ be a finite ADMG, $X$, $Y$ be the cause and effect variables of interest. Denote the subgraph $\mathcal{G}^{-}=\mathcal{G}\left(\left\{ X\right\} \cup\mathrm{An}_{\mathcal{G}}\left(\boldsymbol{Y}^{*}\right)\right)$. The algorithm FDR-TID always has a finite search space and terminates.

    \begin{proof}
        See Appendix~\ref{sec: FDR termination}.
    \end{proof}
\end{lem}
With Lemma~\ref{lemma: FDR-TID termination}, we can conclude that a crude worst-case bound for the time complexity of FDR-TID is that:

\begin{equation}
    T\left(n\right)\leq\sum_{\boldsymbol{S}'\subseteq\left\{ X\right\} \cup\left(\mathrm{An}_{\mathcal{G}^{-}}\left(\boldsymbol{Y}^{*}\right)\setminus\boldsymbol{Y}^{*}\right)}\left(2^{\left|\boldsymbol{Z}_{\boldsymbol{S}'}\right|}-1\right)\kappa T_{\mathrm{op}}\left(\left|\boldsymbol{V}\right|,\left|\boldsymbol{E}\right|\right),
\end{equation}
where $\kappa>0$ is a constant bounding the number of primitive calls per condition check, and $T_{\mathrm{op}}\left(\left|\boldsymbol{V}\right|,\left|\boldsymbol{E}\right|\right)$ is a uniform upper bound on the running time of any single primitive graph operation on an ADMG with $\left|\boldsymbol{V}\right|$ nodes and $\left|\boldsymbol{E}\right|$ arcs. Now we can show the completeness of FDR-TID algorithm.

\begin{thm}[Completeness of FDR-TID]
\label{theorem: complteness of FDR-TID}
    If there exists an FDR triple for a given ADMG $\mathcal{G}\left(\boldsymbol{V}\right)$ and the cause and effect variable pair of interest $\left(X,Y\right)$ , then FDR-TID returns an admissible FDR triple $\left(\boldsymbol{X}^{*},\boldsymbol{Y}^{*},\boldsymbol{M}^{*}\right)$ within finite time.

    \begin{proof}
        Lemma~\ref{lemma: FDR-TID termination} has shown its termination property under finite search steps. The algorithm incrementally enumerates all possible pair $\left(\boldsymbol{S}',\boldsymbol{M}'\right)$ and returns the first admissible FDR triple rather than enumerating all possible ones for the sake of time complexity, which does not affect its completeness.
    \end{proof}
    
\end{thm}
Table~\ref{tab:All-admissible-FDR} shows all admissible FDR triples for ADMGs shown in Figure~\ref{fig:FDR examples} with respect to the cause and effect variable pair of interest $\left(X,Y\right)$. Every admissible FDR triple guarantees that the interventional distribution $p\left(\boldsymbol{y}^{*}\left|do\left(\boldsymbol{x}^{*}\right)\right.\right)$ in the reduced ADMG $\mathcal{G}^{*}$ can be expressed by eq. \eqref{eq:fdr adj formula}. This shows that the FDR criterion together with the FDR-TID algorithm provide a more efficient way to obtain an interpretable, easy-to-estimate and practical interventional distribution expression under a few well-defined criteria, rather than conducting the sophisticated and computationally costly ID algorithm to obtain an equivalent but impractical and hard-to-estimate expression such as shown in eq. \eqref{eq:complex id intv prob}. 

\begin{table}[htbp]
\footnotesize
\begin{centering}
\begin{tabular}{cccc|cccc}
\toprule 
ADMG & $\boldsymbol{X}^{*}$ & $\boldsymbol{Y}^{*}$ & $\boldsymbol{M}^{*}$ & ADMG & $\boldsymbol{X}^{*}$ & $\boldsymbol{Y}^{*}$ & $\boldsymbol{M}^{*}$\tabularnewline
\midrule
\midrule 
\multirow{2}{*}{(a)} & $\left\{ X\right\} $ & \multirow{2}{*}{$\left\{ Y\right\} $} & $\left\{ M\right\} $ & \multirow{2}{*}{(g)} & $\left\{ X\right\} $ & \multirow{2}{*}{$\left\{ Y\right\} $} & $\left\{ M\right\} $,$\left\{ U,M\right\} $\tabularnewline
 & $\left\{ X,V\right\} $ &  & $\left\{ M,Z\right\} $ &  & $\left\{ X,V\right\} $ &  & $\left\{ U\right\} $\tabularnewline
\midrule 
\multirow{2}{*}{(b)} & $\left\{ X\right\} $ & \multirow{2}{*}{$\left\{ Y\right\} $} & \multirow{2}{*}{$\left\{ M\right\} $} & \multirow{2}{*}{(h)} & $\left\{ X\right\} $ & \multirow{2}{*}{$\left\{ Y\right\} $} & $\left\{ M\right\} $,$\left\{ H,M\right\} $,$\left\{ S,M\right\} $,$\left\{ H,S,M\right\} $\tabularnewline
 & $\left\{ X,W\right\} $ &  &  &  & $\left\{ X,H\right\} $ &  & $\left\{ M\right\} $,$\left\{ S,M\right\} $\tabularnewline
\midrule 
\multirow{2}{*}{(c)} & $\left\{ X\right\} $ & \multirow{2}{*}{$\left\{ Y\right\} $} & \multirow{2}{*}{$\left\{ M\right\} $} & \multirow{2}{*}{(i)} & $\left\{ X\right\} $ & \multirow{2}{*}{$\left\{ Y\right\} $} & $\left\{ V,M\right\} $\tabularnewline
 & $\left\{ X,Z\right\} $ &  &  &  & $\left\{ X,W\right\} $ &  & $\left\{ Z,V\right\} $,$\left\{ Z,V,M\right\} $\tabularnewline
\midrule 
\multirow{1}{*}{(d)} & $\left\{ X\right\} $ & \multirow{1}{*}{$\left\{ Y\right\} $} & \multirow{1}{*}{$\left\{ M\right\} $} & (j) & $\left\{ X\right\} $ & $\left\{ Y\right\} $ & $\left\{ U,M\right\} $\tabularnewline
\midrule 
(e) & $\left\{ X\right\} $ & $\left\{ Y\right\} $ & $\left\{ M\right\} $ & (k) & $\left\{ X,U\right\} $ & $\left\{ Y\right\} $ & $\left\{ V,M\right\} $\tabularnewline
\midrule 
(f) & $\left\{ X\right\} $ & $\left\{ Y\right\} $ & $\left\{ M\right\} $,$\left\{ V,M\right\} $ & (l) & $\left\{ X\right\} $ & $\left\{ Y\right\} $ & $\left\{ M\right\} $\tabularnewline
\bottomrule
\end{tabular}
\par\end{centering}
\vspace{1em}
\caption{All admissible FDR triples for ADMGs shown in Figure \ref{fig:FDR examples} with respect to the cause and effect variable pair of interest $\left(X,Y\right)$, where their interventional distributions $p\left(\boldsymbol{y}^{*}\left|do\left(\boldsymbol{x}^{*}\right)\right.\right)$ can be expressed by eq. (\ref{eq:fdr adj formula}). \label{tab:All-admissible-FDR}}
\normalsize
\end{table}

\section{Conclusion}

This paper proposes \emph{front-door reducibility} (FDR), a graphical criterion that extends the reach of front-door adjustment beyond the textbook setting.  First, we formalized FDR through three conditions (FDR1–3) on super-cause,  effect and mediator nodes $(\boldsymbol{X}^{*},\boldsymbol{Y}^{*},\boldsymbol{M}^{*})$.  Second, we proved that FDR is necessary and sufficient for the applicability of an FDR adjustment functional, establishing a graph-level equivalence. Third, we proposed FDR-TID, an exact, terminating algorithm that finds an admissible FDR triple whenever one exists.

These results show that a broad class of ADMGs admit simple, easy-to-estimate adjustment formulas, offering a practical alternative to general ID expressions that are often algebraically complex. Future work includes developing efficient estimators (e.g., doubly robust procedures) tailored to FDR graphs, and integrating FDR-TID into general identification pipelines that preferentially return FDR adjustments when available and fall back to ID only when reduction is impossible.

% \bibliographystyle{unsrtnat}
% \bibliography{refs} 

\newpage
\appendix

\section{Appendix of fundamental causal inference concepts}

In this section, we present some of the most important concepts and definitions commonly used for causal inference. For example, as a fundamental confounding scenario, we give the definition of backdoor criterion as follows:

\begin{defn}[Backdoor criterion \citep{pearl2009causality}]
    \label{def: backdoor}
    Given cause variable $X$, effect variable $Y$ and a set of variables $\boldsymbol{U}$, the causal effect from the cause $X$ to the effect $Y$ is identifiable if:
    
    \begin{enumerate}
        \item No variable in $\boldsymbol{U}$ is a descendant of $X$; and
        \item $\boldsymbol{U}$ blocks every path between $X$ and $Y$ that contains
    an arrow into $X$.
    \end{enumerate}
\end{defn}
Another fundamental concept in causal inference is the \emph{front-door criterion}, which formalizes identification in settings with unobserved confounding between a cause and its effect but where a mediator is observed. Intuitively, the front-door criterion characterizes when a mediator set blocks all directed paths from the cause to the effect, while itself being free of unblocked backdoor paths from the cause and admitting suitable adjustment for its association with the effect. As the core basis of this work, we restate the classical definition below.

\begin{defn}[Front-door criterion \citep{pearl2009causality}]
    \label{def: frontdoor}
    Given cause variable $X$, effect variable $Y$ and a set of variables $\boldsymbol{M}$, the causal effect from the cause $X$ to the effect $Y$ is identifiable if: 
    
    \begin{enumerate}
        \item The mediator variable set $\boldsymbol{M}$ intercepts all directed paths from $X$ to Y; 
        \item There is no unblocked backdoor (see Definition~\ref{def: backdoor}) path from $X$ to $\boldsymbol{M}$; and
        \item All backdoor paths from $\boldsymbol{M}$ to $Y$ are blocked by $X$.
    \end{enumerate}
\end{defn}

And another important concept is the so-called C-component (a.k.a. district), as defined as follows:

\begin{defn}[C-component \citep{pearl2009causality}]
    \label{def: c-component} 
    Let $\mathcal{G}$ be a ADMG such that a subset of its bidirected edges forms a spanning tree over all nodes in $\mathcal{G}$. Then the nodes $\boldsymbol{V}$ in $\mathcal{G}$ form a \emph{C-component} (a.k.a. \emph{district}). If $\mathcal{G}\left(\boldsymbol{V}\right)$ is not a C-component, it can be uniquely partitioned into a set $\mathcal{C}_{\mathcal{G}}$ of subgraphs, each a maximal C-component.
\end{defn}

\section{Proofs}

\subsection{Using \emph{do}-calculus to simplify the interventional distribution expression for Figure~\ref{fig:FDR examples} (f)}
\label{sec: interventional dist derivation Figure 2 (f)}

The example in Figure~\ref{fig:FDR examples}(f) is a special case of our general FDR framework, where the super-nodes reduce to singletons: $\boldsymbol{X}^{*}=\{X\}$, $\boldsymbol{M}^{*}=\{M\}$, and $\boldsymbol{Y}^{*}=\{Y\}$. In the main text, Theorem~\ref{theorem: FDR adjustment} showed that, whenever the FDR conditions (FDR1–3) hold for some triple $(\boldsymbol{X}^{*},\boldsymbol{Y}^{*},\boldsymbol{M}^{*})$, the interventional distribution $p(\boldsymbol{y}^{*}\mid do(\boldsymbol{x}^{*}))$ is given by the FDR adjustment formula \eqref{eq:fdr adj formula}. The derivation below makes this correspondence concrete in the classical front-door setting of Figure~\ref{fig:FDR examples} (f): it applies the three rules of \emph{do}-calculus step by step to show that $p(y\mid do(x))$ reduces exactly to the ordinary front-door adjustment \eqref{eq: ordinary fd adj formula}. In other words, \eqref{eq: interventional dist derivation Figure 2 (f)} can be viewed as the singleton instance of the general derivation used in Theorem~\ref{theorem: FDR adjustment}.

\begin{equation}
\begin{aligned}
    p\left(Y\left|do\left(X\right)\right.\right)= & \int p\left(y\left|do\left(x\right),m\right.\right)p\left(m\left|do\left(x\right)\right.\right)dm\\
    = & \int p\left(y\left|do\left(x\right),do\left(m\right)\right.\right)p\left(m\left|x\right.\right)dm\\
     & \left(\text{Rule 2: }\left(M\perp_{m}Y\left|X\right.\right)_{\mathcal{G}_{\overline{X}\underline{M}}}\text{ and }\left(M\perp_{m}X\right)_{\mathcal{G}_{\underline{X}}}\right)\\
    = & \int p\left(m\left|x\right.\right)p\left(y\left|do\left(m\right)\right.\right)dm\\
     & \left(\text{Rule 3: }\left(Y\perp_{m}X\left|M\right.\right)_{\mathcal{G}_{\overline{X}}}\right)\\
    = & \int p\left(m\left|x\right.\right)\int p\left(y\left|do\left(m\right),x'\right.\right)p\left(x'\left|do\left(m\right)\right.\right)dx'dm\\
    = & \int p\left(m\left|x\right.\right)\int p\left(y\left|m,x'\right.\right)p\left(x'\right)dx'dm\\
     & \left(\text{Rule 2: }\left(Y\perp_{m}M\left|X\right.\right)_{\mathcal{G}_{\underline{M}}}\text{; and Rule 3: }\left(X\perp_{m}M\right)_{\mathcal{G}_{\overline{M}}}\right)
\end{aligned}
\label{eq: interventional dist derivation Figure 2 (f)}
\end{equation}

\begin{figure}[ht]
    \begin{centering}
    \includegraphics[width=0.7\linewidth]{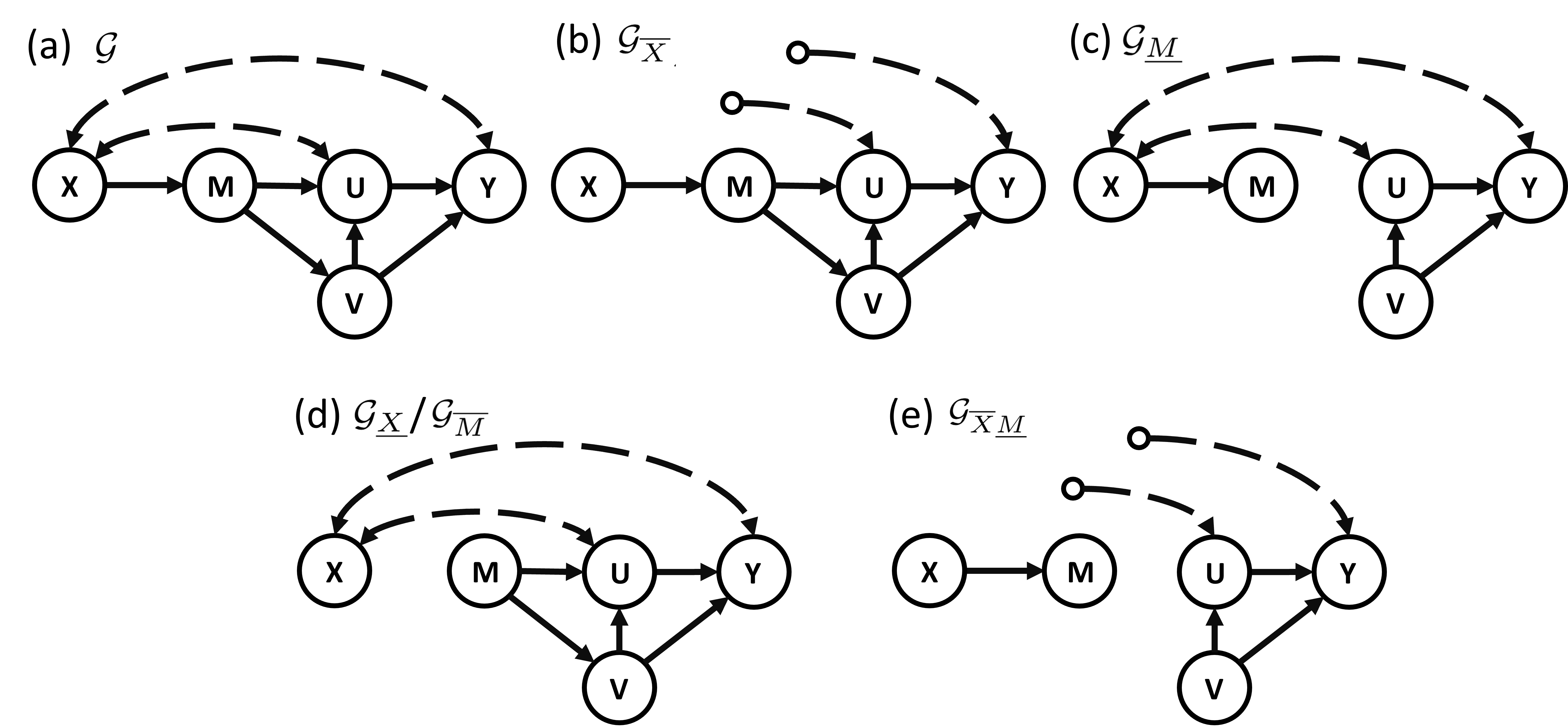}
    \par\end{centering}
    \caption{Example ADMG that satisfies the FDR criterion, together with the modified graphs that appear in the \emph{do}-calculus derivation of eq. \eqref{eq: interventional dist derivation Figure 2 (f)}.}
    \label{fig: interventional dist derivation Figure 2 (f)}
\end{figure}

Figure~\ref{fig: interventional dist derivation Figure 2 (f)} summarizes the sequence of modified graphs that appear in this argument. Each panel corresponds to one of the graphical preconditions required by Rules 2 and 3 of \emph{do}-calculus, and the dashed arcs highlight the backdoor paths that are blocked by the front-door conditions. Thus, the algebraic transformations in eq. \eqref{eq: interventional dist derivation Figure 2 (f)} are not ad hoc: they are exactly the consequences of FDR1–3 in this simple ADMG. The same pattern extends directly to general FDR graphs once variables are aggregated into super-nodes $(\boldsymbol{X}^{*},\boldsymbol{Y}^{*},\boldsymbol{M}^{*})$.

\subsection{Proof for Theorem~\ref{theorem: FDR adjustment}: Front-door reducible adjustment} 
\label{sec: FDR adjustment}
    
 For completeness, we sketch the main steps below using three rules of \emph{do}-calculus to derive the FDR adjustment formula:
        
    \begin{align}
        p\left(\boldsymbol{y}^{*}\left|do\left(\boldsymbol{x}^{*}\right)\right.\right)= & \int p\left(\boldsymbol{y}^{*}\left|do\left(\boldsymbol{x}^{*}\right),\boldsymbol{m}^{*}\right.\right)p\left(\boldsymbol{m}^{*}\left|do\left(\boldsymbol{x}^{*}\right)\right.\right)d\boldsymbol{m}^{*}\label{eq:fdr adj derivation 1}\\
        = & \int p\left(\boldsymbol{y}^{*}\left|do\left(\boldsymbol{x}^{*}\right),do\left(\boldsymbol{m}^{*}\right)\right.\right)p\left(\boldsymbol{m}^{*}\left|do\left(\boldsymbol{x}^{*}\right)\right.\right)d\boldsymbol{m}^{*}\label{eq:fdr adj derivation 2}\\
         & \text{(Rule 2 iteratively \ensuremath{\forall M_{i}\in\boldsymbol{M}^{*}}, \ensuremath{\left(\boldsymbol{Y}^{*}\perp_{m}M_{i}\left|\boldsymbol{X}^{*}\cup\boldsymbol{M}_{-i}^{*}\right.\right)_{\mathcal{G}_{\overline{\boldsymbol{X}^{*}}\underline{M_{i}}}}}, FDR 3)}\nonumber \\
        = & \int p\left(\boldsymbol{y}^{*}\left|do\left(\boldsymbol{m}^{*}\right)\right.\right)p\left(\boldsymbol{m}^{*}\left|\boldsymbol{x}^{*}\right.\right)d\boldsymbol{m}^{*}\label{eq:fdr adj derivation 3}\\
         & \text{(Rule 3 : \ensuremath{\left(\boldsymbol{Y}^{*}\perp_{m}\boldsymbol{X}^{*}\left|\boldsymbol{M}^{*}\right.\right)_{\mathcal{G}_{\overline{\boldsymbol{X}^{*}}}}}, FDR 1; and Rule 2: \ensuremath{\left(\boldsymbol{M}^{*}\perp_{m}\boldsymbol{X}^{*}\right)_{\mathcal{G}_{\underline{\boldsymbol{X}^{*}}}}}, FDR 2)}\nonumber \\
        = & \int p\left(\boldsymbol{m}^{*}\left|\boldsymbol{x}^{*}\right.\right)\int p\left(\boldsymbol{y}^{*}\left|do\left(\boldsymbol{m}^{*}\right),\boldsymbol{x'}^{*}\right.\right)p\left(\boldsymbol{x'}^{*}\left|do\left(\boldsymbol{m}^{*}\right)\right.\right)d\boldsymbol{x'}^{*}d\boldsymbol{m}^{*}\label{eq:fdr adj derivation 4}\\
        = & \int p\left(\boldsymbol{m}^{*}\left|\boldsymbol{x}^{*}\right.\right)\int p\left(\boldsymbol{y}^{*}\left|\boldsymbol{m}^{*},\boldsymbol{x'}^{*}\right.\right)p\left(\boldsymbol{x'}^{*}\right)d\boldsymbol{x'}^{*}d\boldsymbol{m}^{*}\label{eq:fdr adj derivation 5}\\
         & (\text{Rule 2 iteratively }\forall M_{i}\in\boldsymbol{M}^{*},\left(\boldsymbol{Y}^{*}\perp_{m}M_{i}\left|\boldsymbol{X}^{*}\cup\boldsymbol{M}_{-i}^{*}\right.\right)_{\mathcal{G}_{\overline{\boldsymbol{X}^{*}}\underline{M_{i}}}}\text{, FDR 3; }\nonumber \\
         & \text{\text{and Rule3: }\ensuremath{\left(\boldsymbol{X}^{*}\perp_{m}\boldsymbol{M}^{*}\right)_{\mathcal{G}_{\overline{\boldsymbol{M}^{*}}}}}, truncated factorization semantics)}\nonumber 
    \end{align}
            
\subsection{Proof for Theorem~\ref{theorem: equivalence}: Equivalence between FDR adjustment and FDR criterion satisfaction}
\label{sec: FDR criterion and adjustment equivalence}
    To show the equivalence (Statement 1 $\Longleftrightarrow$ Statement 2), we prove both directions:

    \begin{enumerate}
        \item Statement 1 is sufficient for Statement 2: Every step in the derivation eq. \eqref{eq:fdr adj derivation 1}-\eqref{eq:fdr adj derivation 5} that deleting or replacing a $do\left(\cdot\right)$ operator must satisfy the corresponding graphical precondition of the do-calculus rules, e.g., if and only if $\ensuremath{\left(\boldsymbol{M}^{*}\perp_{m}\boldsymbol{X}^{*}\right)_{\mathcal{G}_{\underline{\boldsymbol{X}^{*}}}}}$ to replace $p\left(\boldsymbol{m}^{*}\left|do\left(\boldsymbol{x}^{*}\right)\right.\right)$ with $p\left(\boldsymbol{m}^{*}\left|\boldsymbol{x}^{*}\right.\right)$; otherwise, there exists a model compatible with $\mathcal{G}$ under which that equality fails, proved by Lemma~\ref{lemma: necessity of do-calculus}. Note that the necessity and sufficiency of the manipulation of expressions with $do\left(\cdot\right)$ operators and corresponding graphical preconditions in \emph{do}-calculus rules have been shown in existing literature \citep{huang2012pearl,pearl2009causality}. Let's now consider the key steps in eq. \eqref{eq:fdr adj derivation 1}-\eqref{eq:fdr adj derivation 5}:
    
            \begin{enumerate}
            
                \item In eq. \eqref{eq:fdr adj derivation 3}, we use Rule 2 to replace $p\left(\boldsymbol{m}^{*}\left|do\left(\boldsymbol{x}^{*}\right)\right.\right)$ with $p\left(\boldsymbol{m}^{*}\left|\boldsymbol{x}^{*}\right.\right)$. Then the graph $\mathcal{G}$ must satisfy the graphical precondition $\left(\boldsymbol{M}^{*}\perp_{m}\boldsymbol{X}^{*}\right)_{\mathcal{G}_{\underline{\boldsymbol{X}^{*}}}}$, which corresponds to FDR 2.
                
                \item In eq. \eqref{eq:fdr adj derivation 2} and eq. \eqref{eq:fdr adj derivation 5}, we use Rule 2 to replace $p\left(\boldsymbol{y}^{*}\left|do\left(\boldsymbol{x}^{*}\right),\boldsymbol{m}^{*}\right.\right)$ with $p\left(\boldsymbol{y}^{*}\left|do\left(\boldsymbol{x}^{*}\right),do\left(\boldsymbol{m}^{*}\right)\right.\right)$, and replace $p\left(\boldsymbol{y}^{*}\left|do\left(\boldsymbol{m}^{*}\right),\boldsymbol{x'}^{*}\right.\right)$ with $p\left(\boldsymbol{y}^{*}\left|\boldsymbol{m}^{*},\boldsymbol{x'}^{*}\right.\right)$, respectively. Then the graph $\mathcal{G}$ must satisfy the graphical precondition $\forall M_{i}\in\boldsymbol{M}^{*},\left(\boldsymbol{Y}^{*}\perp_{m}M_{i}\left|\boldsymbol{X}^{*}\cup\boldsymbol{M}_{-i}^{*}\right.\right)_{\mathcal{G}_{\overline{\boldsymbol{X}^{*}}\underline{M_{i}}}}$, which corresponds to FDR 3.
                
                \item In eq. \eqref{eq:fdr adj derivation 3}, we use Rule 3 to replace $p\left(\boldsymbol{y}^{*}\left|do\left(\boldsymbol{x}^{*}\right),do\left(\boldsymbol{m}^{*}\right)\right.\right)$ with $p\left(\boldsymbol{y}^{*}\left|do\left(\boldsymbol{m}^{*}\right)\right.\right)$. Then the graph $\mathcal{G}$ must satisfy the graphical precondition $\ensuremath{\left(\boldsymbol{Y}^{*}\perp_{m}\boldsymbol{X}^{*}\left|\boldsymbol{M}^{*}\right.\right)_{\mathcal{G}_{\overline{\boldsymbol{X}^{*}}}}}$, which is equivalent to the FDR 1, i.e., $\boldsymbol{Y}^{*}\cap\mathrm{De}_{\mathcal{G}\left(\boldsymbol{V}\setminus\boldsymbol{M}\right)}\left(\boldsymbol{X}^{*}\right)=\emptyset$. 
                
                \item Additionally, the step of deleting $do\left(\boldsymbol{m}^{*}\right)$ from $p\left(\boldsymbol{x'}^{*}\left|do\left(\boldsymbol{m}^{*}\right)\right.\right)$ as $p\left(\boldsymbol{x'}^{*}\right)$ in eq. \eqref{eq:fdr adj derivation 5} follows directly from truncated factorization. Since the mediator super node must be the descendants of $\boldsymbol{X}^{*}$, intervening on $\boldsymbol{M}^{*}$ does not affect $\boldsymbol{X}^{*}$. That means no FDR condition is needed for this deletion operation.
            \end{enumerate}
    
        The above proves that if the interventional distribution $p\left(\boldsymbol{y}^{*}\left|do\left(\boldsymbol{x}^{*}\right)\right.\right)$ can be written as eq. \eqref{eq:fdr adj formula} for $\mathcal{G}$, then $\mathcal{G}$ must be front-door reducible to $\mathcal{G}_{\boldsymbol{X}^{*},\boldsymbol{Y}^{*}}^{*}$.
        
        \item Statement 1 is necessary for Statement 2: This has been proved in
        the proof of Theorem~\ref{theorem: FDR adjustment}.
    \end{enumerate}
    With the proved necessity and sufficiency, we prove that the two statements are equivalent.

\subsection{Proof for Proposition~\ref{Prop: effect mini}: Effect minimality}
\label{sec: effect minimality}

    This proposition can be proved by reviewing the criterion shown in~\ref{def: front-door reducible}:
    \begin{enumerate}
        \item The interception of $\boldsymbol{M}^{*}$ is monotone under shrinking the super-effect node, because if $\boldsymbol{M}^{*}$ intercepts all directed paths from $\boldsymbol{X}^{*}$ to $\boldsymbol{Y}^{+}$, it also intercepts those from $\boldsymbol{X}^{*}$ to $\boldsymbol{Y}^{*}$.
        
        \item If FDR 2 holds for triple $\left(\boldsymbol{X}^{*},\boldsymbol{Y}^{+},\boldsymbol{M}^{*}\right)$, then it must satisfy for $\left(\boldsymbol{X}^{*},\boldsymbol{Y}^{*},\boldsymbol{M}^{*}\right)$, because it is independent of the super-effect $\boldsymbol{Y}^{*}$.
        
        \item For all $M_{i}\in\boldsymbol{M}^{*}$, if $\left(\boldsymbol{Y}^{+}\perp_{m}M_{i}\left|\boldsymbol{X}^{*}\cup\boldsymbol{M}_{-i}\right.\right)_{\mathcal{G}_{\overline{\boldsymbol{X}^{*}}\underline{M_{i}}}}$, then $\left(Y\perp_{m}M_{i}\left|\boldsymbol{X}^{*}\cup\boldsymbol{M}_{-i}\right.\right)_{\mathcal{G}_{\overline{\boldsymbol{X}^{*}}\underline{M_{i}}}}$ since $Y\in\boldsymbol{Y}^{+}$.
    \end{enumerate}

\subsection{Proof for Lemma~\ref{lemma:Completeness of X}: Completeness of the potential super-cause family \texorpdfstring{$\boldsymbol{\mathcal{X}}$}{X}}
\label{sec: Completeness of the potential super-cause family}
    By Definition~\ref{def: FDR triple}, the potential super-cause family is $\boldsymbol{\mathcal{X}}=\left\{ \boldsymbol{S}\ensuremath{\subseteq\left\{ X\right\} \cup\left(\mathrm{An}_{\mathcal{G}}\left(\boldsymbol{Y}^{*}\right)\setminus\boldsymbol{Y}^{*}\right)}:X\in\boldsymbol{S}\right\} $. Since $\boldsymbol{X}'\subseteq\left\{ X\right\} \cup\left(\mathrm{An}\left(\boldsymbol{Y}^{*}\right)\setminus\boldsymbol{Y}^{*}\right)$ and $X\in\boldsymbol{X}'$, we have $\boldsymbol{X}'\in\boldsymbol{\mathcal{X}}$. Let $\boldsymbol{S}_{0}=\boldsymbol{X}^{+}\setminus\boldsymbol{X}'$ to represent the shrinking set. Then, by construction, $\boldsymbol{S}_{0}\cap\mathrm{An}\left(\boldsymbol{Y}^{*}\right)=\emptyset$. We show that we can safely replace $\boldsymbol{X}^{+}$ by $\boldsymbol{X}'$ in the FDR triple without violating the FDR criterion, possibly after shrinking $\boldsymbol{M}^{+}$ to a some $\boldsymbol{M}^{'}\subseteq\boldsymbol{M}^{+}$:
    
    \begin{enumerate}
        \item \textbf{FDR 1:} If there exists a directed path $\pi_{\boldsymbol{X}'\rightarrow\boldsymbol{Y}^{*}}$ from some node in $\boldsymbol{X}'$ to some node in $\boldsymbol{Y}^{*}$ that is not intercepted by $\boldsymbol{M}^{+}$, then it is also a directed path $\pi_{\boldsymbol{X}^{+}\rightarrow\boldsymbol{Y}^{*}}$ without interception by $\boldsymbol{M}^{+}$, contradicting the FDR condition for $\left(\boldsymbol{X}^{+},\boldsymbol{Y}^{*},\boldsymbol{M}^{+}\right)$. Moreover, by construction we have $\boldsymbol{S}_{0}\cap\mathrm{An}(\boldsymbol{Y}^{*})=\emptyset$, so there is no directed path $\pi_{\boldsymbol{S}_{0}\rightarrow\boldsymbol{Y}^{*}}$, which means removing $\boldsymbol{S}_{0}$ cannot create a new directed path from $\boldsymbol{X}'$ to $\boldsymbol{Y}^{*}$ that avoids $\boldsymbol{M}^{+}$. This proves that shrinking $\boldsymbol{X}^{+}$ to $\boldsymbol{X}'$ does not violate FDR 1. \item \textbf{FDR 2:} Let $\mathcal{G}_{\boldsymbol{\underline{X}}^{+}}$ and $\mathcal{G}_{\boldsymbol{\underline{X}}'}$ be the graphs obtained from G by deleting all outgoing arcs from $\boldsymbol{X}^{+}$ and $\boldsymbol{X}'$, respectively. Then $\mathcal{G}_{\boldsymbol{\underline{X}}^{+}}$ can be obtained from $\mathcal{G}_{\boldsymbol{\underline{X}}'}$ by additionally deleting all outgoing edges from $\boldsymbol{S}_{0}$. FDR 2 for the FDR triple $\left(\boldsymbol{X}^{+},\boldsymbol{Y}^{*},\boldsymbol{M}^{+}\right)$ implies that $\left(\boldsymbol{X}^{+}\perp_{m}\boldsymbol{M}^{+}\right)_{\mathcal{G}_{\boldsymbol{\underline{X}}^{+}}}$. 
        
        Suppose that FDR 2 fails for $\left(\boldsymbol{X}',\boldsymbol{Y}^{*},\boldsymbol{M}^{+}\right)$. Then there exists an \emph{m-connected} path between some $X_{i}'\in\boldsymbol{X}'$ and some $M_{i}^{+}\in\boldsymbol{M}^{+}$ in $\mathcal{G}_{\boldsymbol{\underline{X}}'}$. Since the only difference between $\mathcal{G}_{\boldsymbol{\underline{X}}^{+}}$ and $\mathcal{G}_{\boldsymbol{\underline{X}}'}$ is the presence of outgoing arcs from $\boldsymbol{S}_{0}$, any such new \emph{m-connected} path must use at least one node in $\boldsymbol{S}_{0}$ as a non-collider with a directed path toward $M_{0}^{+}$. Hence, that node lies in $\mathrm{An}\left(M_{0}^{+}\right)$.
        
        Moreover, for FDR we may \emph{w.l.o.g.} restrict $\boldsymbol{M}^{+}$ to mediators on directed paths $\pi_{\boldsymbol{X}^{+}\rightarrow\boldsymbol{Y}^{*}}$, since any mediator outside all directed paths from $\boldsymbol{X}^{+}$ to $\boldsymbol{Y}^{*}$ can be removed without affecting FDR 1-3. Denote this subset $\boldsymbol{M}^{'}\subseteq\boldsymbol{M}^{+}$, then $\boldsymbol{M}^{'}\subseteq\mathrm{An}\left(\boldsymbol{Y}^{*}\right)$, hence $\mathrm{An}\left(\boldsymbol{M}'\right)\subseteq\mathrm{An}\left(\boldsymbol{Y}^{*}\right)$. Thus $\forall S_{i}\in\boldsymbol{S}_{0}$, such that $S_{i}\in\mathrm{An}\left(\boldsymbol{Y}^{*}\right)$, contradicting $\boldsymbol{S}_{0}\cap\mathrm{An}(\boldsymbol{Y}^{*})=\emptyset$. Therefore, FDR 2 is preserved when replacing $\boldsymbol{X}^{+}$ by $\boldsymbol{X}'$ (with the shrunk super-mediator $\boldsymbol{M}^{'}$) 
        
        \item \textbf{FDR 3:} In the modified graph $\mathcal{G}_{\overline{\boldsymbol{X}^{+}}\underline{M_{i}^{+}}}$ , FDR 3 requires $\forall M_{i}^{+}\in\boldsymbol{M}^{+}$, such that $\left(\boldsymbol{Y}^{*}\perp_{m}M_{i}^{+}\left|\boldsymbol{X}^{+}\cup\boldsymbol{M}_{-i}^{+}\right.\right)_{\mathcal{G}_{\overline{\boldsymbol{X}^{+}}\underline{M_{i}^{+}}}}$. If we instead delete only the coming arcs into $\boldsymbol{X}'$, we have the modified graph $\mathcal{G}_{\overline{\boldsymbol{X}'}\underline{M_{i}^{+}}}$ which can be obtained from $\mathcal{G}_{\overline{\boldsymbol{X}^{+}}\underline{M_{i}^{+}}}$ by adding some incoming arcs into nodes $S_{i}\in\boldsymbol{S}_{0}$ . Thus the only possible way to create a new \emph{m-connected} path between $M_{i}^{+}$ and $\boldsymbol{Y}^{*}$(conditional on $\boldsymbol{X}'\cup\boldsymbol{M}_{-i}^{+}$) in $\mathcal{G}_{\overline{\boldsymbol{X}'}\underline{M_{i}^{+}}}$ is to use $S_{i}$ as a non-collider on that path.
        
        But if $S_{i}$ appears as a non-collider on a conditionally \emph{m-connected} path between $M_{i}^{+}$ and $\boldsymbol{Y}^{*}$, then $S_{i}\in\mathrm{An}\left(\boldsymbol{Y}^{*}\right)\cup\mathrm{An}\left(M_{i}^{+}\right)$. Using the same restricted super-mediator set $\boldsymbol{M}^{'}\subseteq\mathrm{An}\left(\boldsymbol{Y}^{*}\right)$ as above, we conclude again that $S_{i}\in\mathrm{An}\left(\boldsymbol{Y}^{*}\right)$, contradicting $\boldsymbol{S}_{0}\cap\mathrm{An}\left(\boldsymbol{Y}^{*}\right)=\emptyset$. Hence FDR 3 is also preserved when shrinking $\boldsymbol{X}^{+}$ to $\boldsymbol{X}'$ with the shrunk super-mediator $\boldsymbol{M}^{'}$.
    \end{enumerate}

\subsection{Proof for Lemma~\ref{lemma: Completeness of Z_S}: Completenesss of the candidate mediator region \texorpdfstring{$\boldsymbol{Z}\left(\boldsymbol{S}'\right)$}{Z}}
\label{sec: Completenesss of the candidate mediator region}

    Start from any feasible $\boldsymbol{M}^{+} \subseteq \boldsymbol{Z}\left(\boldsymbol{S}'\right)$, the pruning involves three aspects without breaking the FDR criterion:
    \begin{enumerate}
    
        \item Restriction to the interior region $\boldsymbol{I}=\mathrm{An}_{\mathcal{G}}\left(\boldsymbol{Y}^{*}\right)\cap\mathrm{De}_{\mathcal{G}}\left(\boldsymbol{S}'\right)\setminus\left(\boldsymbol{S}'\cup\boldsymbol{Y}^{*}\right)$: $\boldsymbol{I}$ is then the set of all interior nodes on directed paths $\pi_{\boldsymbol{S}'\rightarrow\boldsymbol{Y}^{*}}$. If $\boldsymbol{M}^{+}$ is a feasible super-mediator that satisfies FDR criterion, then for any $M_{i}^{+}\in\boldsymbol{M}^{+}$ such that $M_{i}^{+}\notin\boldsymbol{I}$, $M_{i}^{+}$ cannot on any directed path of $\pi_{\boldsymbol{S}'\rightarrow\boldsymbol{Y}^{*}}$. So, any node $V_{i}\notin\boldsymbol{I}$ can be safely excluded.
        
        \item Removal of the first bidirected-arc-connected neighbours of $\boldsymbol{S}'$ ($\mathrm{Nb}_{\mathcal{G}}^{\leftrightarrow}\left(\boldsymbol{S}'\right)$): If there exists $M_{i}^{+}\in\boldsymbol{M}^{+}\cap\mathrm{Nb}_{\mathcal{G}}^{\leftrightarrow}\left(\boldsymbol{S}'\right)$, then there must exist a m-connected path $\left(M_{i}^{+}\leftrightarrow\boldsymbol{S}'\right)$ in the modified ADMG $\mathcal{G}_{\boldsymbol{\underline{S}}'}$, contradicting FDR 2. So, removing $\mathrm{Nb}_{\mathcal{G}}^{\leftrightarrow}\left(\boldsymbol{S}'\right)$ is safe.
        
        \item Removal of the first bidirected-arc-connected neighbours of $\boldsymbol{Y}^{*}$ ($\mathrm{Nb}_{\mathcal{G}}^{\leftrightarrow}\left(\boldsymbol{Y}^{*}\right)$): If there exists $M_{i}^{+}\in\boldsymbol{M}^{+}\cap\mathrm{Nb}_{\mathcal{G}}^{\leftrightarrow}\left(\boldsymbol{Y}^{*}\right)$, then there must exist a m-connected path $\left(M_{i}^{+}\leftrightarrow\boldsymbol{Y}^{*}\right)$ in the modified ADMG $\mathcal{G}_{\overline{\boldsymbol{X}^{+}}\underline{M_{i}^{+}}}$ regardless of conditioning on $\boldsymbol{S}'\cup\boldsymbol{M}_{-i}$, contradicting FDR 3. So, removing $\mathrm{Nb}_{\mathcal{G}}^{\leftrightarrow}\left(\boldsymbol{Y}^{*}\right)$ is safe.
    \end{enumerate}

\subsection{Proof for Lemma~\ref{lemma: FDR-TID termination}: FDR-TID finite search space and termination}
\label{sec: FDR termination}
    The search space for $\boldsymbol{X}^{*}$ and $\boldsymbol{M}^{*}$ are finite, and each FDR condition check in the FDR-TID algorithm must terminate in a finite round:

    \begin{enumerate}
    \item Finite $\boldsymbol{X}^{*}$-search. Note that $\boldsymbol{X}^{*}$ takes the first admissible supersets $\boldsymbol{S}'\ni X$ if $\mathcal{G}$ is FDR. FDR-TID enumerates $\boldsymbol{S}'$ inside $\mathrm{An}_{\mathcal{G}}\left(\boldsymbol{Y}^{*}\right)\setminus\boldsymbol{Y}^{*}$, where the size of search space is $2^{\left|\mathrm{An}_{\mathcal{G}}\left(\boldsymbol{Y}^{*}\right)\setminus\left(\boldsymbol{Y}^{*}\cup\left\{ X\right\} \right)\right|}$, hence finite.
    
    \item Finite $\boldsymbol{M}^{*}$-search per $\boldsymbol{S}'$. For each $\boldsymbol{S}'$, the candidate mediator region $\boldsymbol{Z}_{\boldsymbol{S}'}$ is a finite set as shown at line 5 (also in \eqref{eq:candidate mediator region}). FDR-TID enumerates nonempty $\boldsymbol{M}'\subseteq\boldsymbol{Z}_{\boldsymbol{S}'}$, whose size of search space is $2^{\left|\boldsymbol{Z}_{\boldsymbol{S}'}\right|}-1<\infty$.
    
    \item Each FDR condition check must terminate. For every pair of $\left(\boldsymbol{S}',\boldsymbol{M}'\right)$, the three-condition checks are done by a finite number of graph operations, e.g., edge deletions, m-separation queries on an ADMG. Each such graph operation run in time polynomial in $\left|\boldsymbol{V}\right|+\left|\boldsymbol{E}\right|$, where $\left|\boldsymbol{E}\right|$ is the total number of arcs in an ADMG.
    
    \end{enumerate}
    Therefore, FDR-TID always terminates no later than the first admissible FDR triple is found.

\section{Appendix of \emph{do}-calculus rules}

The \emph{do}-calculus \citep{pearl1995causal} consists of three rules that can be used to transform expressions involving $do\left(\cdot\right)$ operators into other expressions of this type, whenever certain graphical preconditions are satisfied in the causal diagram (ADMG) $\mathcal{G}$. We give the definition of the three rules of \emph{do}-calculus as follows:

\begin{defn}[Three rules of \textbf{\emph{do}-calculus} \citep{pearl1995causal}]
    \label{def: do-calculus rules}
    The following three rules are valid for every interventional distribution compatible with any ADMG $\mathcal{G}$:
    \begin{itemize}
    
        \item Rule 1 (Insertion/deletion of observations):
        \begin{equation}
            p\left(y\left|do\left(x\right),z,w\right.\right)=p\left(y\left|do\left(x\right),w\right.\right)\text{ if }\left(Y\perp_{m}Z\left|X,W\right.\right)_{\mathcal{G}_{\overline{X}}}
        \end{equation}
        
        \item Rule 2 (Intervention/observation exchange):
        \begin{equation}
            p\left(y\left|do\left(x\right),do\left(z\right),w\right.\right)=p\left(y\left|do\left(x\right),z,w\right.\right)\text{ if }\left(Y\perp_{m}Z\left|X,W\right.\right)_{\mathcal{G}_{\overline{X}\underline{Z}}}
        \end{equation}
        
        \item Rule 3 (Insertion/deletion of Interventions):
        \begin{equation}
            p\left(y\left|do\left(x\right),do\left(z\right),w\right.\right)=p\left(y\left|do\left(x\right),w\right.\right)\text{ if }\left(Y\perp_{m}Z\left|X,W\right.\right)_{\mathcal{G}_{\overline{X}\overline{Z\left(W\right)}}},
        \end{equation}
        where $Z\left(W\right)$ is the set of $Z$-nodes that are not ancestors of any $W$-node in $\mathcal{G}_{\overline{X}}$.
    \end{itemize}
\end{defn}
As a preparation to prove the equivalence between FDR adjustment and FDR criterion satisfaction so as in Theorem~\ref{theorem: equivalence}, we need to show a necessity lemma for these \emph{do}-calculus rules asserting that the corresponding graphical preconditions are required for the equalities to hold for all models compatible with $\mathcal{G}$.

\begin{lem}[The necessity of do-calculus rules]
    \label{lemma: necessity of do-calculus}
    For disjoint $Y,Z,X,W$ if:
    \begin{enumerate}
        \item $p\left(y\left|do\left(x\right),z,w\right.\right)=p\left(y\left|do\left(x\right),w\right.\right)$ holds for all models compatible with $\mathcal{G},$ then $\left(Y\perp_{m}Z\left|X,W\right.\right)_{\mathcal{G}_{\overline{X}}}$.
        
        \item $p\left(y\left|do\left(x\right),do\left(z\right),w\right.\right)=p\left(y\left|do\left(x\right),z,w\right.\right)$ holds for all models compatible with $\mathcal{G},$ then $\left(Y\perp_{m}Z\left|X,W\right.\right)_{\mathcal{G}_{\overline{X}\underline{Z}}}$.
        
        \item $p\left(y\left|do\left(x\right),do\left(z\right),w\right.\right)=p\left(y\left|do\left(x\right),w\right.\right)$ holds for all models compatible with $\mathcal{G},$ then $\left(Y\perp_{m}Z\left|X,W\right.\right)_{\mathcal{G}_{\overline{X}\overline{Z\left(W\right)}}}$.
    \end{enumerate}
    \begin{proof}
        The proof is set up with two standard facts:
        \begin{enumerate}
        
            \item The existence of faithful distributions for ADMGs: if $Z$ and $W$ are not m-separated by $U$ in a modified ADMG $\mathcal{G}'$ (e.g. $\mathcal{G}_{\overline{X}}$, $\mathcal{G}_{\overline{X}\underline{Z}}$ and $\mathcal{G}_{\overline{X}\overline{Z\left(W\right)}}$) , there exists a distribution $Q$ compatible with $\mathcal{G}'$ such that $Z\not\perp_{m}W\mid U$ under $Q$.
            
            \item The truncated factorization: for any $Q$ compatible with a modified ADMG $\mathcal{G}'\left(\boldsymbol{V}\right)$, there exists a structural causal model (SCM) $\mathcal{M}$ compatible with $\mathcal{G}$ whose post-intervention distribution under $do\left(S\right)$ is $Q=p_{\mathcal{M}}\left(\boldsymbol{V}\left|do\left(S\right)\right.\right)$.
        \end{enumerate}
        
        And so we are able to argue by contraposition using the two facts above for the three rules:
        \begin{enumerate}
        
        \item Rule 1: If $\left(Y\not\perp_{m}Z\left|X,W\right.\right)_{\mathcal{G}_{\overline{X}}}$, pick $Q$ compatible with $\mathcal{G}_{\overline{X}}$ where $\left(Y\not\perp_{m}Z\left|X,W\right.\right)_{\mathcal{G}_{\overline{X}}}$, and lift to a SCM $\mathcal{M}$ compatible with $\mathcal{G}'$, then $p_{\mathcal{M}}\left(y\left|do\left(x\right),z,w\right.\right)\neq p_{\mathcal{M}}\left(y\left|do\left(x\right),w\right.\right)$. This contradicts the universal validity, which implies $\left(Y\perp_{m}Z\left|X,W\right.\right)_{\mathcal{G}_{\overline{X}}}$.
        
        \item Rule 2: If $\left(Y\not\perp_{m}Z\left|X,W\right.\right)_{\mathcal{G}_{\overline{X}\underline{Z}}}$, pick $Q$ compatible with $\mathcal{G}_{\overline{X}\underline{Z}}$, and lift to a SCM $\mathcal{M}$ compatible with $\mathcal{G}'$, then $p\left(y\left|do\left(x\right),do\left(z\right),w\right.\right)\neq p\left(y\left|do\left(x\right),z,w\right.\right)$. That also contradicts the universal validity, which implies $\left(Y\perp_{m}Z\left|X,W\right.\right)_{\mathcal{G}_{\overline{X}\underline{Z}}}$.
        
        \item Rule 3: If $\left(Y\not\perp_{m}Z\left|X,W\right.\right)_{\mathcal{G}_{\overline{X}\overline{Z\left(W\right)}}}$, pick $Q$ compatible with $\mathcal{G}_{\overline{X}\overline{Z\left(W\right)}}$, and lift to a SCM $\mathcal{M},$then $p\left(y\left|do\left(x\right),do\left(z\right),w\right.\right)\neq p\left(y\left|do\left(x\right),w\right.\right)$. That again contradicts the universal validity, which implies $\left(Y\perp_{m}Z\left|X,W\right.\right)_{\mathcal{G}_{\overline{X}\overline{Z\left(W\right)}}}$.
        \end{enumerate}
    \end{proof}
\end{lem}

\end{document}